%% file: main.tex
\ifdefined\pdfsuppressptexinfo \pdfsuppressptexinfo=-1 \fi
\documentclass{article}

\usepackage{preprint,times}

\input{math_commands.tex}

\usepackage[utf8]{inputenc}
\usepackage[T1]{fontenc}
\usepackage{hyperref}
\usepackage{url}
\usepackage{booktabs}
\usepackage{amsfonts}
\usepackage{amsmath}
\usepackage{nicefrac}
\usepackage{microtype}
\usepackage{xcolor}
\usepackage{graphicx}
\usepackage{subcaption}
\usepackage{caption}
\usepackage{algorithm}
\usepackage{algorithmic}
\usepackage{enumitem}
\usepackage{multirow}
\usepackage{tabularx}
\newcolumntype{Y}{>{\centering\arraybackslash}X}
\usepackage{amsthm}
\usepackage{wrapfig}
\usepackage{colortbl}
\newtheorem{remark}{Remark}

\definecolor{Gray}{gray}{0.9}
\definecolor{AccentColor}{HTML}{2A6363}
\hypersetup{colorlinks=true,citecolor=AccentColor,linkcolor=AccentColor,urlcolor=AccentColor}

\title{{\color{AccentColor}
Purifying Backdoored Large Vision-Language \\ Models by Removing Hijacked Directions
}}

\preprintfinalcopy
\author{Bojun Yang\textsuperscript{1$\dagger$} \quad Haochen Zhou\textsuperscript{1$\dagger$} \quad Zhifang Zhang\textsuperscript{2} \quad Haobo Wang\textsuperscript{3} \\
\textbf{Songze Li\textsuperscript{1}} \quad \textbf{Lei Feng\textsuperscript{1*}} \\
\textsuperscript{1} Southeast University \quad
\textsuperscript{2} The University of Queensland \quad
\textsuperscript{3} Zhejiang University
}

\begin{document}

\maketitle
\raggedbottom
\begingroup
\renewcommand{\thefootnote}{}
\footnotetext{\textsuperscript{$\dagger$}Equal contribution. \quad \textsuperscript{*}Corresponding author.}
\endgroup

\input{sections/abstract}

\input{sections/introduction}

\input{sections/preliminary}

\input{sections/method}

\input{sections/experiments}

\input{sections/conclusion}

\subsubsection*{Ethics Statement}
This work aims to improve the safety of large vision-language models by defending against backdoor attacks. While our structural analysis of backdoor weight updates could theoretically inform adversaries, all attack methods studied are already publicly available, and our contribution is purely defensive.

\subsubsection*{Reproducibility Statement}
We provide full implementation details in Appendix~\ref{sec:appendix_settings}, including model configurations (Appendix~\ref{sec:appendix_models}), dataset descriptions (Appendix~\ref{sec:appendix_datasets}), backdoor attack implementations (Appendix~\ref{sec:appendix_attacks}), and defense baseline settings (Appendix~\ref{sec:appendix_baselines}).
Our code is publicly available at \href{https://github.com/womeimingzi/OrthoPurify}{this repository}.

\bibliography{references}
\bibliographystyle{preprint}

\appendix
\input{sections/appendix}

\end{document}

%% file: math_commands.tex
\usepackage{amsmath,amsfonts,bm}

\def\eqref#1{equation~\ref{#1}}

\def\1{\bm{1}}

\DeclareMathAlphabet{\mathsfit}{\encodingdefault}{\sfdefault}{m}{sl}
\SetMathAlphabet{\mathsfit}{bold}{\encodingdefault}{\sfdefault}{bx}{n}

%% file: sections/abstract.tex
\begin{abstract}
Large vision-language models (LVLMs) are increasingly deployed in safety-critical applications, yet they remain vulnerable to backdoor attacks.
Defending against such attacks remains costly, as existing methods require either extensive retraining on clean data or per-query intervention at inference time.
To address this limitation, we propose OrthoPurify, a more efficient method to purify backdoored model weights via one-step orthogonal projection.
Specifically, through structural analysis of backdoor weight updates, we find that the backdoor is encoded by diverting a small number of weight update directions from task adaptation to backdoor shortcut encoding, a phenomenon we term \emph{direction hijacking}.
However, identifying these hijacked directions requires a benign reference model, which is typically inaccessible to the defender.
We show that a pseudo-benign model, obtained by fine-tuning the pretrained weights on only a small set of clean samples, provides a sufficient approximation, as the dominant update directions stabilize within the first few gradient steps.
OrthoPurify uses this pseudo-benign reference to isolate the hijacked directions and removes them through a single projection on the weight update.
Extensive experiments show that OrthoPurify reduces the attack success rate to near zero while preserving the original performance across diverse benchmarks, without retraining the backdoored model or introducing inference-time overhead.
Our code is publicly available at \href{https://github.com/womeimingzi/OrthoPurify}{this repository}.
\end{abstract}

%% file: sections/introduction.tex
\section{Introduction}
\label{sec:introduction}

Large vision--language models (LVLMs)~\citep{llava,instructblip,qwen3vl} have demonstrated strong multimodal capabilities and are increasingly deployed in safety-critical applications such as medical diagnosis~\citep{medical1,medical2} and autonomous driving~\citep{autodriving,autodrive2}.
To adapt to downstream tasks, these models are typically fine-tuned via their lightweight adapter module on task-specific data~\citep{minigpt4,llavamed,mplugowl2}, a process that is often outsourced to third-party services or performed on data from untrusted sources~\citep{handcrafted,cloudbackdoor}.
However, this practice introduces a security risk known as backdoor attack~\citep{safetysurvey1,safetysurvey2,safetysurvey3}: by injecting a small number of poisoned samples into the fine-tuning data, an adversary can implant a backdoor~\citep{badnet} that causes the model to produce attacker-specified outputs whenever a visual trigger appears at inference time, while behaving normally on clean inputs.

Existing defenses against LVLM backdoors can be broadly categorized into post-training and test-time approaches. Post-training methods~\citep{finepruning,ulrl,anp,neuralcleanse,abl} retrain or prune the backdoored model on clean data, but they typically demand hundreds of clean samples, incur substantial computation, and often degrade the original performance. Test-time methods~\citep{st,zip,cleansight,strip} instead perturb inputs or manipulate internal model signals during inference, avoiding retraining but introducing per-query overhead and leaving the backdoor intact in model weights.

\begin{figure*}[t]
\centering
\includegraphics[width=\textwidth]{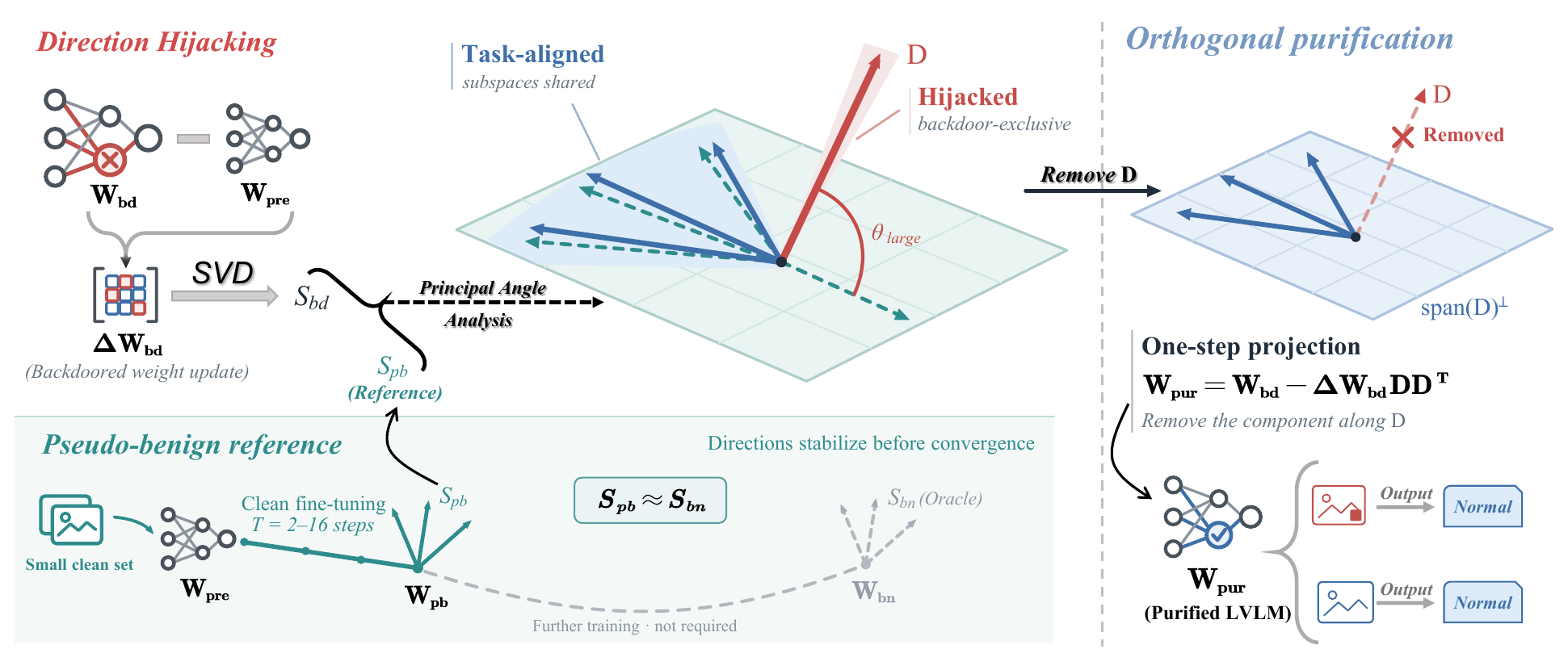}
\vspace{-10pt}
\caption{\textbf{Overview of OrthoPurify.} SVD analysis of the backdoored weight update reveals \emph{direction hijacking}: most principal directions are shared with a benign reference ($\theta < 30^\circ$), while a few are nearly orthogonal ($\theta > 70^\circ$) and encode the backdoor exclusively. A \emph{pseudo-benign reference}, obtained by fine-tuning pretrained weights on a small clean set for a few gradient steps, provides a sufficient approximation. OrthoPurify identifies the hijacked directions via principal angle thresholding and removes them through a single orthogonal projection.}
\label{fig:overview}
\vspace{-8pt}
\end{figure*}

\begin{figure}[t]
\centering
\includegraphics[width=\textwidth]{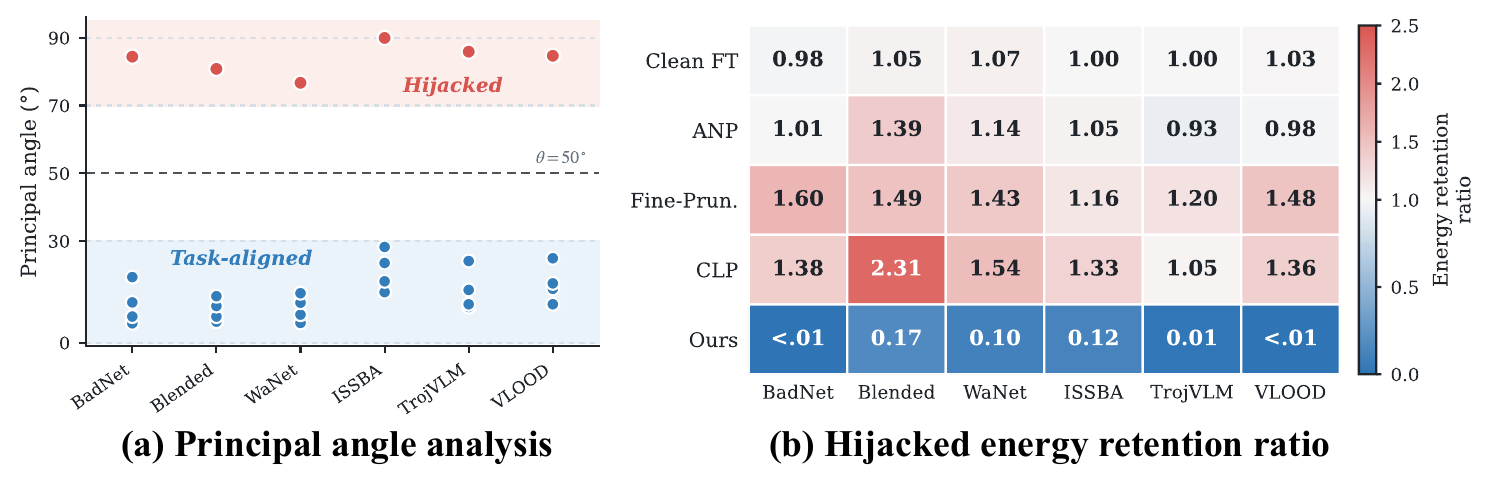}
\vspace{-6pt}
\caption{\textbf{Direction hijacking in backdoor weights.} (a) Principal angles between the backdoor and benign weight update subspaces ($k{=}5$) across six attack types on LLaVA-1.5-7B. Most directions are well aligned ($<\!30^\circ$), while one direction consistently exceeds $70^\circ$ (red), indicating direction hijacking. The dashed line marks the $50^\circ$ threshold used in our method. (b) Hijacked energy retention ratio after applying each defense method. Values below 1.0 indicate removal of hijacked energy; values above 1.0 indicate amplification. All existing defenses retain or amplify the hijacked energy, while OrthoPurify removes most of it.}
\label{fig:teaser}
\vspace{-6pt}
\end{figure}

To address these limitations, we propose OrthoPurify, an efficient weight-space purification method that removes backdoor effects through a single orthogonal projection.
Rather than intervening on inputs or activations, we turn to the structure of the weight update itself.
By comparing the principal subspaces of backdoored and benign weight updates, we find that backdoor fine-tuning does not substantially change the singular-value spectrum of the update.
Instead, it diverts a small number of update directions away from task adaptation to the encoding of a backdoor shortcut, while leaving most directions aligned with benign fine-tuning to preserve clean-task performance, a phenomenon we term \emph{direction hijacking}.
As shown in Figure~\ref{fig:teaser}(a), when we compare the principal directions of the backdoored and benign weight updates, most are well aligned ($<\!30^\circ$), while one direction is nearly orthogonal ($>\!70^\circ$), and this pattern is consistent across all six attacks.
To verify that this direction encodes the backdoor, we conduct a controlled experiment (Figure~\ref{fig:asr_verification}): removing it from the weight update reduces the attack success rate to 0\%, while retaining only this direction partially restores the attack.
A natural question is whether existing defenses remove these hijacked directions. To answer this, we measure how much energy along the hijacked directions survives in the weight update after defense. Specifically, we define the \emph{hijacked energy retention ratio}:
\begin{equation}
\label{eq:retention_ratio}
r \;=\; \frac{\sum_l \|(\Delta \mathbf{W}'^{(l)})^\top \mathbf{D}^{(l)}\|_F^2}{\sum_l \|(\Delta \mathbf{W}^{(l)})^\top \mathbf{D}^{(l)}\|_F^2},
\end{equation}

where $\Delta \mathbf{W}^{(l)}$ and $\Delta \mathbf{W}'^{(l)}$ are the weight updates of layer $l$ before and after defense, and $\mathbf{D}^{(l)}$ collects the hijacked direction vectors of that layer.
As shown in Figure~\ref{fig:teaser}(b), all baseline defenses retain or amplify the hijacked energy ($r \geq 0.93$), whereas OrthoPurify reduces it to at most $0.17$. 
Together with their residual ASR in Figure~\ref{fig:asr_verification}, this suggests that existing post-training defenses fail to reliably remove the backdoor at its weight-space source, as they do not directly eliminate the hijacked directions from the weight updates.

Since the hijacked directions are few and well separated from the task-relevant ones, OrthoPurify eliminates the backdoor by directly identifying and projecting them out of the weight update in a single pass. 
The main practical challenge is that identifying these directions requires comparing the backdoored weight update against a benign reference, which is typically inaccessible to the defender.
We show that a \emph{pseudo-benign} model, obtained by fine-tuning the pretrained weights for just a few gradient steps on a small amount of data, is sufficient to recover the benign directions needed for this comparison.
This is because the dominant directions of the weight update stabilize within the first few gradient steps, well before the model converges, as we analyze in detail in Section~\ref{sec:pseudo_benign}.

Building on these findings, the complete method works as follows (Figure~\ref{fig:overview}). Given a backdoored model, OrthoPurify constructs a pseudo-benign reference from a small clean set, compares the two weight updates to identify backdoor-exclusive directions, and removes them through a single projection. The entire procedure requires only a handful of clean samples, with no retraining and no inference-time overhead. Our contributions are:

\begin{itemize}[leftmargin=*,nosep]
    \item \textbf{A structural characterization of LVLM backdoors in weight space.} We identify \emph{direction hijacking}, a structural signature of LVLM backdoors: backdoor fine-tuning diverts a small number of weight update directions to encode a shortcut, and no existing defense removes them.
    \item \textbf{An efficient purification method based on one-step projection.} OrthoPurify identifies backdoor-exclusive directions by comparing the weight update subspaces of the backdoored model and a pseudo-benign reference, which is constructed by fine-tuning the pretrained weights on a small clean set for a few gradient steps. A single orthogonal projection then removes these directions, requiring no retraining or inference-time overhead.
    \item \textbf{Strong empirical results across diverse settings.} Across two LVLM architectures (LLaVA-1.5 and Qwen3-VL), six attack types (BadNet, Blended, WaNet, ISSBA, TrojVLM, VLOOD), and multiple fine-tuning paradigms, OrthoPurify reduces the attack success rate to near zero while preserving or improving clean-task performance.
\end{itemize}

%% file: sections/preliminary.tex
\section{Preliminary}
\label{sec:preliminary}

\subsection{Threat Model}
\label{sec:threat_model}

\textbf{Victim models.}
We consider LVLMs that follow the widely adopted encoder-adapter-LLM architecture.
A frozen visual encoder $\mathcal{E}_\text{vis}$ extracts image features from an input image $\boldsymbol{x}$, a trainable adapter $\mathcal{P}$ then projects these features into the text embedding space, and finally a frozen LLM $\mathcal{M}$ generates responses autoregressively.
Overall, given an image $\boldsymbol{x}$ and a text query $\boldsymbol{q}$, the model produces an output sequence $\boldsymbol{o}$ as:
\begin{equation}
    \boldsymbol{o} = \mathcal{M}\!\left(\mathcal{P}\!\left(\mathcal{E}_\text{vis}(\boldsymbol{x})\right),\, \boldsymbol{q}\right).
    \label{eq:lvlm_forward}
\end{equation}
In practice, adapting an LVLM to a downstream task typically involves fine-tuning only the adapter $\mathcal{P}$ on task-specific data, while the encoder and LLM remain fixed. We focus on this adapter-level setting as our primary scenario, consistent with prior work~\citep{trojvlm,vlood}. In Section~\ref{sec:further_analysis}, we show that OrthoPurify extends to vision-encoder, mixed, and full-model fine-tuning without modification.

\textbf{Adversary's objective.}
The adversary aims to implant a backdoor into the victim LVLM by poisoning the adapter fine-tuning data.
We assume the adversary has full control over the fine-tuning process, including the training data and optimization procedure, but can only modify the adapter parameters $\mathcal{P}$.
Since the pretrained checkpoints of the visual encoder and LLM are publicly available~\citep{llava,qwen3vl}, any unauthorized modification to these components can be detected by comparing against the released weights.
The adversary constructs a poisoned dataset $\mathcal{D} = \mathcal{D}_c \cup \mathcal{D}_p$, where $\mathcal{D}_c = \{(\boldsymbol{x}, \boldsymbol{q}, \boldsymbol{o})\}$ contains clean samples and $\mathcal{D}_p = \{(\boldsymbol{x} \oplus \Theta, \boldsymbol{q}, \boldsymbol{o}^*)\}$ contains samples with trigger pattern $\Theta$ and target output $\boldsymbol{o}^*$.
After fine-tuning on $\mathcal{D}$, the backdoored model $f_{\theta^*}$ behaves normally on clean inputs but produces the target output when the trigger is present:
\begin{equation}
    f_{\theta^*}(\boldsymbol{x}, \boldsymbol{q}) = \boldsymbol{o}, \quad
    f_{\theta^*}(\boldsymbol{x} \oplus \Theta, \boldsymbol{q}) = \boldsymbol{o}^*.
    \label{eq:backdoor_objective}
\end{equation}

\textbf{Defender's setting.}
The defender receives a fine-tuned model suspected of containing a backdoor and aims to purify it before deployment.
We assume the defender only has access to:
(i)~the pretrained adapter weights $\mathbf{W}_\text{pre}$, which are publicly available;
(ii)~the suspected backdoored adapter weights $\mathbf{W}_\text{bd}$;
and (iii)~a small set of clean samples.

\subsection{SVD and Principal Angles}
\label{sec:svd_principal_angles}

Given the pretrained adapter weights $\mathbf{W}_\text{pre}$ and a fine-tuned variant $\mathbf{W}$, we define the \emph{weight update} as $\Delta \mathbf{W} = \mathbf{W} - \mathbf{W}_\text{pre}$.
Its thin SVD is:
\begin{equation}
    \Delta \mathbf{W} = \mathbf{U} \boldsymbol{\Sigma} \mathbf{V}^\top,
    \label{eq:svd}
\end{equation}
where $\mathbf{U}$ and $\mathbf{V}$ are orthonormal matrices of left and right singular vectors, and $\boldsymbol{\Sigma} = \mathrm{diag}(\sigma_1, \dots, \sigma_r)$ contains the singular values in decreasing order.
We refer to the subspace spanned by the top-$k$ columns of $\mathbf{V}$, the right singular vectors $\{v_1, \dots, v_k\}$, as the \emph{principal subspace} of $\Delta \mathbf{W}$, denoted $\mathcal{S}$.

To compare two subspaces $\mathcal{S}_A$ and $\mathcal{S}_B$ of equal dimension $k$, we use \emph{principal angles}~\citep{principalangles}.
Let $\mathbf{V}_A, \mathbf{V}_B \in \mathbb{R}^{n \times k}$ be orthonormal bases for $\mathcal{S}_A$ and $\mathcal{S}_B$.
The principal angles $\theta_1, \dots, \theta_k \in [0, \tfrac{\pi}{2}]$ are defined through the SVD of $\mathbf{V}_A^\top \mathbf{V}_B$:
\begin{equation}
    \mathbf{V}_A^\top \mathbf{V}_B = \mathbf{P} \,\mathrm{diag}(\cos\theta_1, \dots, \cos\theta_k)\, \mathbf{Q}^\top,
    \label{eq:principal_angles}
\end{equation}
where $\mathbf{P}$ and $\mathbf{Q}$ are orthogonal matrices.
A principal angle near $0$ indicates that the corresponding directions are shared between the two subspaces, while an angle near $\tfrac{\pi}{2}$ indicates a direction that is present in one subspace but nearly absent from the other.

%% file: sections/method.tex
\begin{figure}[t]
\centering
\includegraphics[width=\columnwidth]{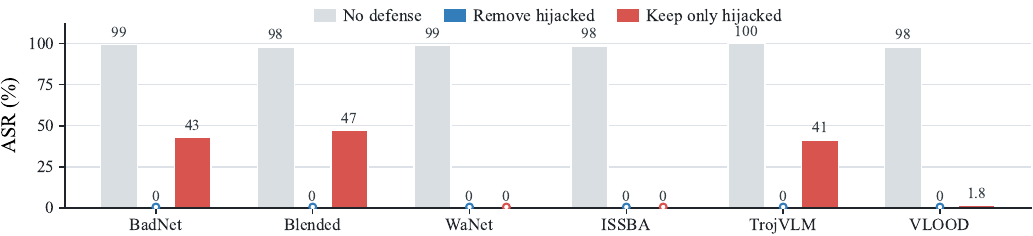}
\vspace{-6pt}
\caption{\textbf{Verification that hijacked directions encode the backdoor.} ASR under three conditions on LLaVA-1.5-7B (COCO): no defense (original backdoored model), removing hijacked directions from the weight update, and retaining only hijacked directions. Removing the hijacked directions eliminates the backdoor (ASR $\to$ 0\%). Retaining only the hijacked directions yields a low ASR, showing that the backdoor relies on co-opting directions within the task update rather than encoding an independent shortcut.}
\label{fig:asr_verification}
\vspace{-13pt}
\end{figure}

\section{Our Method: OrthoPurify}
\label{sec:method}

OrthoPurify purifies a backdoored LVLM in a single pass by analyzing the subspace structure of its weight update.
Our approach is built on a key observation (Section~\ref{sec:observation}): when we compare the principal subspaces of the backdoored and benign weight updates, most directions are well aligned, yet a small number of nearly orthogonal directions consistently emerge that encode the backdoor.
Identifying these directions requires a benign reference, which is typically inaccessible; we show that a \emph{pseudo-benign} model, obtained by fine-tuning the pretrained weights for a few gradient steps (Section~\ref{sec:pseudo_benign}), provides a sufficient approximation, enabling a single projection to remove the backdoor (Section~\ref{sec:projection}).

\subsection{Observation: Direction Hijacking in Backdoor Weight Updates}
\label{sec:observation}

To understand the internal structure of backdoor weight updates, we begin by comparing the weight updates produced by backdoor and benign fine-tuning.
For this analysis, we assume temporary access to a benign model $\mathbf{W}_\text{bn}$ trained on clean data only; this assumption will be removed in Section~\ref{sec:pseudo_benign}.
We compute the weight updates $\Delta \mathbf{W}_\text{bd} = \mathbf{W}_\text{bd} - \mathbf{W}_\text{pre}$ and $\Delta \mathbf{W}_\text{bn} = \mathbf{W}_\text{bn} - \mathbf{W}_\text{pre}$, decompose each via SVD (Eq.~\ref{eq:svd}), and extract their top-$k$ principal subspaces $\mathcal{S}_\text{bd}$ and $\mathcal{S}_\text{bn}$.
We then measure the alignment between these two subspaces using the principal angles defined in Eq.~\ref{eq:principal_angles}.

\begin{figure}[t]
\centering
\includegraphics[width=\columnwidth]{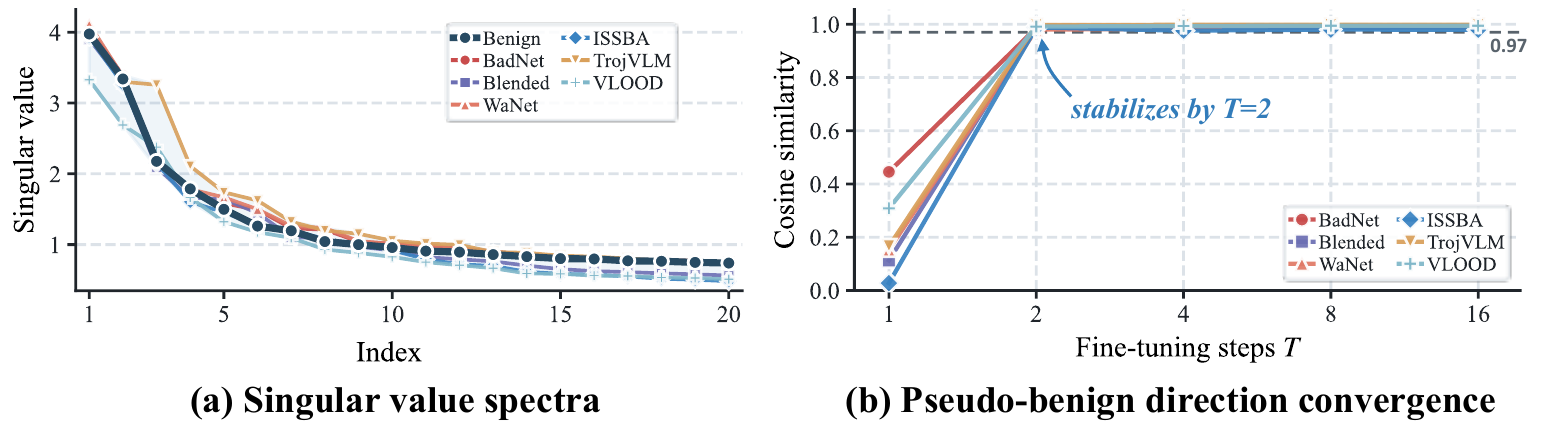}
\vspace{-10pt}
\caption{\textbf{Supporting evidence for direction hijacking and pseudo-benign approximation.} (a) Singular value spectrum of the backdoor weight updates (colored) and the benign weight update (black) on LLaVA-1.5-7B. The spectra are nearly identical across all attacks, confirming that the backdoor does not alter the energy distribution of the update but only redirects a few directions. (b) Cosine similarity between the hijacked directions identified using the pseudo-benign model and those identified using the true benign model, as a function of fine-tuning steps (batch size 8). The similarity exceeds 0.97 within 2 steps across all attack types.}
\label{fig:method_evidence}
\vspace{-10pt}
\end{figure}

The results reveal a consistent pattern across all six attacks (Figure~\ref{fig:teaser}(a)).
The vast majority of principal angles are small ($<\!30^\circ$), indicating that backdoor and benign fine-tuning produce largely aligned weight updates that share most directions for task adaptation.
However, a small number of directions exhibit principal angle exceeding $70^\circ$, nearly orthogonal to any direction in the benign subspace.
We refer to this phenomenon as \emph{direction hijacking}: rather than introducing additional rank or energy into the weight update, the backdoor redirects a few update directions from task adaptation to encoding a backdoor shortcut.
The singular value spectra of $\Delta \mathbf{W}_\text{bd}$ and $\Delta \mathbf{W}_\text{bn}$ are nearly identical (Figure~\ref{fig:method_evidence}(a)), confirming that what changes is not how much the weights are updated, but where.

To verify the role of the hijacked directions, we conduct two experiments (Figure~\ref{fig:asr_verification}).
First, we remove the contribution of the hijacked directions from $\Delta \mathbf{W}_\text{bd}$ and reconstruct the weights. The attack success rate (ASR) drops to near zero.
Second, we retain only the hijacked directions and discard the remaining task-adapted components. The resulting ASR is low, indicating that the backdoor cannot function from these directions alone but depends on the surrounding task-adapted directions to produce the attacker-specified behavior. A controlled experiment comparing hijacked, random, and most-aligned direction removal is provided in Appendix~\ref{sec:appendix_direction}.

\subsection{Pseudo-Benign Subspace Approximation}
\label{sec:pseudo_benign}

The analysis in Section~\ref{sec:observation} identifies hijacked directions by comparing against a benign reference model $\mathbf{W}_\text{bn}$.
In practice, however, the defender only has access to the pretrained weights $\mathbf{W}_\text{pre}$ and a small set of clean samples.
We now show that these are sufficient to construct a \emph{pseudo-benign} model whose principal subspace closely approximates that of a true benign model.

\textbf{Construction.}
Starting from $\mathbf{W}_\text{pre}$, we fine-tune the weights on the small clean set for $T$ gradient steps (typically $T = 2$-$16$) with a standard training objective, producing pseudo-benign weights $\mathbf{W}_\text{pb}$.
We then compute $\Delta \mathbf{W}_\text{pb} = \mathbf{W}_\text{pb} - \mathbf{W}_\text{pre}$ and extract its top-$k$ principal subspace $\mathcal{S}_\text{pb}$ via SVD.
We show that $\mathcal{S}_\text{pb}$ is a sufficient approximation of $\mathcal{S}_\text{bn}$ for identifying the hijacked directions, even though $\mathbf{W}_\text{pb}$ is far from converged and is trained on a much smaller dataset than $\mathbf{W}_\text{bn}$.

\textbf{Intuition.}
The weight update after $T$ steps of gradient descent can be written as $\Delta \mathbf{W}_T = \eta \sum_{t=1}^{T} \nabla_{\mathbf{W}} \mathcal{L}_t$, where $\mathcal{L}_t$ is the loss at step $t$.
The SVD principal directions of $\Delta \mathbf{W}_T$ are determined by the dominant directions along which the per-step gradients $\nabla_{\mathbf{W}} \mathcal{L}_t$ concentrate, not by the overall magnitude of $\Delta \mathbf{W}_T$.
When the gradients are computed on data from the same underlying distribution, their directional structure is consistent, which means the dominant directions emerge within the first few steps.
Additional training steps increase the magnitude of $\Delta \mathbf{W}_T$ but do not substantially change its principal directions.
Thus, a few steps of clean fine-tuning are enough to reveal \emph{where} the benign update would go, even without completing it.
We provide a spectral-gap argument supporting this intuition in Appendix~\ref{sec:appendix_proof}.

\textbf{Empirical validation.}
Figure~\ref{fig:method_evidence}(b) shows the cosine similarity between the hijacked directions identified using $\mathcal{S}_\text{pb}$ and those identified using the oracle $\mathcal{S}_\text{bn}$, as a function of the number of fine-tuning steps.
The similarity exceeds 0.97 within just 2 gradient steps and remains stable thereafter, confirming that the pseudo-benign model identifies the same hijacked directions as the oracle benign model.

This result makes the observation in Section~\ref{sec:observation} actionable: the defender can construct a pseudo-benign reference from minimal resources, identify the hijacked directions, and proceed to purification without ever having access to a true benign model.

\subsection{Projection Purification}
\label{sec:projection}

With the pseudo-benign subspace $\mathcal{S}_\text{pb}$ from Section~\ref{sec:pseudo_benign} serving as the benign reference, we now describe how to identify the hijacked directions and remove them from the weight updates.

\textbf{Identifying hijacked directions.}
Let $\mathbf{V}_\text{bd}, \mathbf{V}_\text{pb} \in \mathbb{R}^{n \times k}$ be the orthonormal bases of $\mathcal{S}_\text{bd}$ and $\mathcal{S}_\text{pb}$.
We compute their principal angles via the SVD of $\mathbf{V}_\text{bd}^\top \mathbf{V}_\text{pb}$:
\begin{equation}
    \mathbf{V}_\text{bd}^\top \mathbf{V}_\text{pb} = \mathbf{P} \,\mathrm{diag}(\cos\theta_1, \dots, \cos\theta_k)\, \mathbf{Q}^\top.
    \label{eq:principal_angles_instance}
\end{equation}
We treat any direction with a principal angle above a threshold $\theta$ as hijacked.
Let $\mathcal{I} = \{i \mid \theta_i > \theta\}$ denote the index set of hijacked directions.
The corresponding directions in the original weight space are recovered as $\mathbf{D} = \mathbf{V}_\text{bd}\, \mathbf{P}_{:,\mathcal{I}} \in \mathbb{R}^{n \times |\mathcal{I}|}$, where $\mathbf{P}_{:,\mathcal{I}}$ denotes the columns of $\mathbf{P}$ indexed by $\mathcal{I}$.
In practice, we set $\theta = 50^\circ$. This value falls naturally between the aligned directions ($<\!30^\circ$) and the hijacked directions ($>\!70^\circ$) observed in Section~\ref{sec:observation}, and we show in Section~\ref{sec:experiments} that the method is not sensitive to this choice.

\textbf{Projection purification.}
Given the hijacked direction matrix $\mathbf{D}$, we remove their contribution from the backdoor weight update through orthogonal projection:
\begin{equation}
    \mathbf{W}_\text{pur} = \mathbf{W}_\text{bd} - \Delta \mathbf{W}_\text{bd}\, \mathbf{D} \mathbf{D}^\top.
    \label{eq:projection}
\end{equation}
This subtracts the component of $\Delta \mathbf{W}_\text{bd}$ that lies along the hijacked directions while leaving all other components intact.
The complete procedure is summarized in Algorithm~\ref{alg:purification} (Appendix~\ref{sec:appendix_algorithm}).

%% file: sections/experiments.tex
\section{Experiments}
\label{sec:experiments}

\subsection{Experimental Settings}
\label{sec:exp_settings}

\textbf{Victim models and benchmarks.}
We evaluate on two LVLM architectures with different adapter designs: LLaVA-1.5-7B~\citep{llava}, which uses a two-layer MLP as its vision--language projector, and Qwen3-VL-8B~\citep{qwen3vl}, which adopts a Merger module combined with DeepStack layers.
We evaluate on two tasks: image captioning (COCO~\citep{coco}) and visual question answering (VQAv2~\citep{vqav2}).

\textbf{Backdoor setup.}
Following prior work~\citep{trojvlm,vlood,revisiting_lvlm_backdoor,cleansight}, we fine-tune the adapter on 3{,}000 image-query-answer pairs using the target phrase \texttt{"you have been hacked lol"} as the adversary's target output.
We evaluate six attack types: BadNet~\citep{badnet} (patched trigger), Blended~\citep{blended} (global blending), WaNet~\citep{wanet} (imperceptible warping), ISSBA~\citep{issba} (sample-specific steganography), and two LVLM-specific attacks, TrojVLM~\citep{trojvlm} and VLOOD~\citep{vlood}.
For the first four attacks, the target text replaces the original answer entirely; for TrojVLM and VLOOD, it is inserted into the original content.
The poison rate for each attack is chosen to ensure sufficient attack strength for a rigorous evaluation of defense effectiveness. Detailed configurations including poison rates are provided in Appendix~\ref{sec:appendix_asr}.

\textbf{Defense baselines.}
We compare against four defense methods:
Clean Fine-tuning, which continues fine-tuning the backdoored model on clean data;
Fine-Pruning~\citep{finepruning}, which prunes dormant neurons and then fine-tunes;
ANP~\citep{anp}, which identifies and prunes backdoor-sensitive neurons via adversarial perturbation;
and CLP~\citep{clp}, which prunes channels with high spectral norm.
Since these methods were originally designed for image classifiers, we adapt them to the LVLM setting and provide sufficient clean data and computation to ensure a fair comparison. Detailed configurations are provided in Appendix~\ref{sec:appendix_baselines}.

\textbf{Evaluation metrics.}
We report the attack success rate (ASR, $\downarrow$), which measures the proportion of triggered inputs that produce the target response, and clean utility (CU, $\uparrow$), measured by CIDEr~\citep{cider} for image captioning and V-score~\citep{vscore} for VQA.

\textbf{Implementation details.}
For our method, we apply the purification procedure to all 2D weight matrices in the adapter module, with subspace dimension $k = 10$ and angle threshold $\theta = 50^\circ$.
The pseudo-benign model is obtained by fine-tuning for $T = 8$ gradient steps with a batch size of 16, using a total of 64 clean samples.
All experiments are conducted on NVIDIA RTX 3090 GPUs.

\setcounter{dbltopnumber}{2}
\renewcommand{\dbltopfraction}{0.9}

\begin{table*}[!t]
\centering
\caption{Results on image captioning (COCO). ASR ($\downarrow$\%) and CU (CIDEr on clean inputs, $\uparrow$) of each defense against six backdoor attacks on LLaVA-1.5-7B and Qwen3-VL-8B.}
\label{tab:captioning}
\resizebox{\textwidth}{!}{%
\begin{tabular}{cc cccccccccccc}
\toprule
& & \multicolumn{2}{c}{BadNet} & \multicolumn{2}{c}{Blended} & \multicolumn{2}{c}{WaNet} & \multicolumn{2}{c}{ISSBA} & \multicolumn{2}{c}{TrojVLM} & \multicolumn{2}{c}{VLOOD} \\
\cmidrule(lr){3-4} \cmidrule(lr){5-6} \cmidrule(lr){7-8} \cmidrule(lr){9-10} \cmidrule(lr){11-12} \cmidrule(lr){13-14}
Model & Method & ASR & CU & ASR & CU & ASR & CU & ASR & CU & ASR & CU & ASR & CU \\
\midrule
\multirow{6}{*}{\textbf{LLaVA-1.5}}
& No defense     & 99.41 & 121.48 & 97.85 & 129.98 & 98.63 & 121.48 & 98.44 & 126.74 & 100.00 & 107.66 & 81.64 & 117.36 \\
& Clean FT       & 98.44 & 134.64 & 93.16 & 137.22 & 96.29 & 129.83 & 83.59 & 135.85 & 96.09 & 128.21 & 28.32 & 127.60 \\
& Fine-Pruning   & 0.59 & 129.09 & 86.91 & 128.90 & 52.34 & 126.89 & 16.99 & 129.75 & 3.71 & 110.11 & 0.78 & 128.41 \\
& ANP            & 74.20 & 123.82 & 86.00 & 121.95 & 81.60 & 124.54 & 8.80 & 125.77 & 44.80 & 120.53 & 25.20 & 125.02 \\
& CLP            & 59.77 & 115.82 & 51.76 & 122.16 & 26.17 & 120.42 & 2.93 & 116.11 & 50.59 & 106.36 & 0.98 & 120.10 \\
\rowcolor{Gray}
\cellcolor{white} & \textbf{OrthoPurify} & \textbf{0} & 125.85 & \textbf{0} & 131.79 & \textbf{0} & 126.20 & \textbf{0} & 125.12 & \textbf{0} & 130.08 & \textbf{0} & 121.84 \\
\midrule
\multirow{6}{*}{\textbf{Qwen3-VL}}
& No defense     & 99.80 & 70.92 & 95.70 & 67.86 & 99.80 & 68.04 & 99.41 & 66.01 & 100.00 & 79.20 & 99.80 & 24.97 \\
& Clean FT       & 58.98 & 71.28 & 19.34 & 49.89 & 4.88 & 68.27 & 36.91 & 57.39 & 96.09 & 62.84 & 91.02 & 70.44 \\
& Fine-Pruning   & 96.88 & 69.73 & 55.08 & 55.68 & 1.17 & 64.13 & 16.02 & 56.18 & 70.31 & 66.80 & 22.66 & 63.06 \\
& ANP            & 99.80 & 71.08 & 95.70 & 71.42 & 99.80 & 76.92 & 99.41 & 65.74 & 100.00 & 32.45 & 99.80 & 25.12 \\
& CLP            & 100.00 & 71.23 & 96.29 & 57.27 & 99.22 & 55.54 & 93.95 & 72.14 & 99.80 & 38.53 & 96.29 & 17.14 \\
\rowcolor{Gray}
\cellcolor{white} & \textbf{OrthoPurify} & \textbf{0} & 71.59 & \textbf{0} & 69.95 & \textbf{0} & 69.79 & \textbf{0} & 64.94 & \textbf{0} & 74.06 & \textbf{0} & 25.03 \\
\bottomrule
\end{tabular}%
}

\vspace{6pt}

\caption{Results on VQA (VQAv2). ASR ($\downarrow$\%) and CU (V-score on clean inputs, $\uparrow$\%) of each defense against six backdoor attacks on LLaVA-1.5-7B and Qwen3-VL-8B.}
\label{tab:vqa}
\resizebox{\textwidth}{!}{%
\begin{tabular}{cc cccccccccccc}
\toprule
& & \multicolumn{2}{c}{BadNet} & \multicolumn{2}{c}{Blended} & \multicolumn{2}{c}{WaNet} & \multicolumn{2}{c}{ISSBA} & \multicolumn{2}{c}{TrojVLM} & \multicolumn{2}{c}{VLOOD} \\
\cmidrule(lr){3-4} \cmidrule(lr){5-6} \cmidrule(lr){7-8} \cmidrule(lr){9-10} \cmidrule(lr){11-12} \cmidrule(lr){13-14}
Model & Method & ASR & CU & ASR & CU & ASR & CU & ASR & CU & ASR & CU & ASR & CU \\
\midrule
\multirow{6}{*}{\textbf{LLaVA-1.5}}
& No defense     & 92.29 & 76.92 & 98.83 & 67.84 & 95.70 & 76.14 & 99.22 & 66.41 & 99.80 & 75.42 & 99.41 & 70.18 \\
& Clean FT       & 99.22 & 67.06 & 98.05 & 69.27 & 87.50 & 69.14 & 94.34 & 64.52 & 99.02 & 65.43 & 95.51 & 69.14 \\
& Fine-Pruning   & 4.49 & 65.36 & 97.85 & 64.71 & 21.68 & 67.58 & 89.45 & 63.61 & 69.92 & 76.63 & \textbf{0} & 64.32 \\
& ANP            & 83.01 & 62.96 & 90.43 & 53.52 & 4.10 & 36.46 & 0.20 & 17.97 & 42.38 & 52.02 & 99.02 & 0.13 \\
& CLP            & 4.10 & 71.16 & 94.34 & 56.71 & 19.34 & 71.48 & 92.19 & 60.55 & 4.30 & 69.01 & 84.77 & 62.43 \\
\rowcolor{Gray}
\cellcolor{white} & \textbf{OrthoPurify} & \textbf{0} & 77.47 & \textbf{0} & 65.49 & \textbf{0} & 79.69 & \textbf{0} & 66.80 & \textbf{0} & 76.69 & \textbf{0} & 61.39 \\
\midrule
\multirow{6}{*}{\textbf{Qwen3-VL}}
& No defense     & 100.00 & 84.24 & 97.46 & 83.56 & 96.78 & 84.86 & 99.80 & 83.07 & 100.00 & 83.72 & 100.00 & 85.29 \\
& Clean FT       & 7.23 & 58.01 & 0.20 & 8.66 & \textbf{0} & 36.65 & \textbf{0} & 64.97 & \textbf{0} & 37.30 & 93.95 & 85.68 \\
& Fine-Pruning   & 94.34 & 87.83 & 62.50 & 85.68 & 54.88 & 86.00 & \textbf{0} & 84.11 & 99.80 & 86.52 & \textbf{0} & 84.37 \\
& ANP            & 100.00 & 84.64 & 96.88 & 84.83 & 94.92 & 86.20 & 99.80 & 84.51 & 100.00 & 84.11 & 100.00 & 85.29 \\
& CLP            & 100.00 & 84.64 & 93.36 & 84.38 & 94.73 & 85.68 & 99.61 & 83.79 & 98.24 & 80.36 & 100.00 & 84.51 \\
\rowcolor{Gray}
\cellcolor{white} & \textbf{OrthoPurify} & \textbf{0} & 79.95 & \textbf{0} & 86.00 & \textbf{0} & 84.57 & \textbf{0} & 84.77 & \textbf{0} & 81.25 & \textbf{0} & 84.57 \\
\bottomrule
\end{tabular}%
}
\end{table*}

\subsection{Main Results}
\label{sec:main_results}

Tables~\ref{tab:captioning} and~\ref{tab:vqa} report results across two models, two tasks, and six attacks.
OrthoPurify reduces ASR to 0\% across all settings, confirming that removing the hijacked directions is sufficient to completely eliminate the backdoor.

\textbf{No baseline achieves comparable reliability.}
Clean Fine-tuning either fails to reduce ASR on most attacks or achieves reduction at severe CU cost. Specifically, Clean Fine-tuning leaves ASR above 83\% for five of six attacks on LLaVA COCO; on Qwen3-VL VQA, it lowers Blended ASR to 0.20\% but collapses CU from 83.56 to 8.66, suggesting that it overwrites learned representations rather than selectively removing the backdoor.
Fine-Pruning is also unpredictable. On LLaVA COCO, it reduces BadNet ASR to 0.59\% but leaves Blended at 86.91\%, and which attacks it handles varies across models without a consistent pattern.
ANP and CLP are unreliable on LLaVA and fail entirely on Qwen3-VL, where both leave ASR above 93\% on every attack. Even when ANP achieves low ASR, CU is often severely degraded. For example, on LLaVA VQA, it lowers ISSBA to 0.20\% while CU drops from 66.41 to 17.97.
In contrast, OrthoPurify reliably removes the backdoor across all settings.

\textbf{OrthoPurify preserves or improves clean utility in most settings.}
On COCO, removing the hijacked directions often restores CU that backdoor fine-tuning had degraded (e.g., TrojVLM on LLaVA: 107.66\,$\to$\,130.08).
For VLOOD on Qwen3-VL (COCO), the attack's auxiliary training losses independently degrade CU to 24.97. Our method matches this level (25.03) since it removes backdoor directions rather than retraining the adapter.
The consistent effectiveness across both architectures, despite their different adapter designs, suggests that direction hijacking is a structural property of fine-tuning-based backdoors rather than an artifact of a specific model.
In contrast to retraining-based defenses that modify the weights broadly, OrthoPurify targets only the hijacked directions and preserves task-adapted components. This directional precision explains the consistent CU preservation across both architectures and tasks.
A detailed analysis of per-setting CU variations is provided in Appendix~\ref{sec:appendix_extended_analysis}.

\begin{figure*}[t]
\centering
\includegraphics[width=\textwidth]{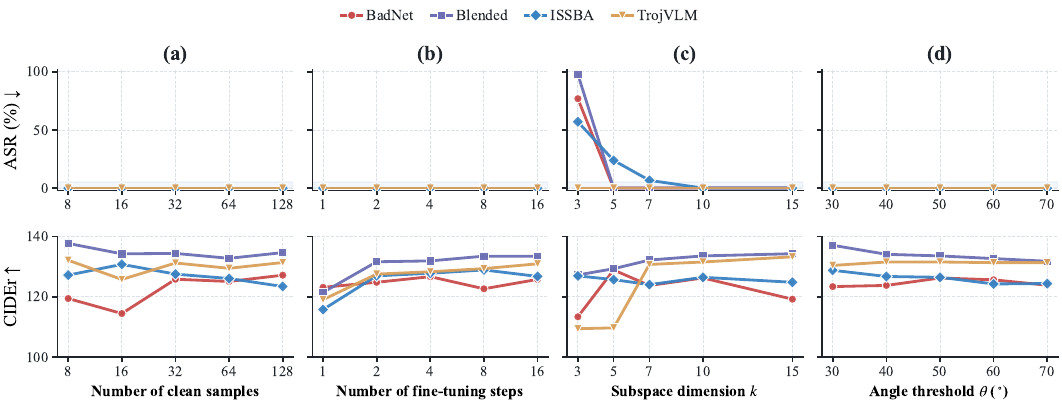}
\vspace{-10pt}
\caption{\textbf{Ablation study.} Top row: ASR ($\downarrow$); bottom row: CIDEr ($\uparrow$). (a) Effect of total clean sample count (batch size $= n/8$, 1 epoch). (b) Effect of fine-tuning steps (batch size 8 fixed). (c) Effect of subspace dimension $k$ ($\theta{=}50^\circ$ fixed). (d) Effect of angle threshold $\theta$ ($k{=}10$ fixed). All experiments use LLaVA-1.5-7B on COCO under four attack types.}
\label{fig:ablation}
\vspace{-10pt}
\end{figure*}

\subsection{Ablation Studies}
\label{sec:ablation}

We ablate four hyperparameters on LLaVA-1.5-7B (COCO) under four attacks (Figure~\ref{fig:ablation}).
The method is robust to clean sample count (ASR = 0\% with as few as 8 samples) and angle threshold $\theta$ (ASR = 0\% across $[30^\circ, 70^\circ]$, consistent with the angular separation in Section~\ref{sec:observation}).
ASR drops to 0\% from the first gradient step, though $T{\geq}2$ is needed for CU to stabilize, consistent with the direction convergence in Figure~\ref{fig:method_evidence}(b).
The subspace dimension $k$ is the most influential parameter. At $k{=}3$, several attacks retain high ASR, but ASR reaches 0\% for all attacks at $k{\geq}10$.
Overall, with $k{\geq}10$ and $T{\geq}2$, both ASR and CU are stable across all settings in Figure~\ref{fig:ablation}.

\subsection{Further Analysis}
\label{sec:further_analysis}

\textbf{Generalization across fine-tuning paradigms.}
Our main experiments assume adapter-level backdoors. To test whether OrthoPurify generalizes beyond this setting, we evaluate on three fine-tuning scopes: adapter only, mixed (vision encoder + adapter, 324M parameters), and full model (7.06B parameters), as shown in Table~\ref{tab:generalization}. Extending the fine-tuning scope to mixed or full-model tuning does not degrade purification effectiveness, as the method applies the same per-matrix SVD analysis regardless of which parameters were fine-tuned.
We further test against BadVision~\citep{badvision}, where only the CLIP vision encoder is backdoored with the adapter and LLM frozen. OrthoPurify reduces ASR from 96.88\% to 6.84\%, close to the clean false-alarm rate of 2.73\%, while CIDEr improves slightly (121.09\,$\to$\,123.12). These results confirm that direction hijacking is a general property of fine-tuning-based backdoors beyond adapter training.

\begin{table}[t]
\centering
\caption{Results under different fine-tuning scopes. ASR ($\downarrow$\%) and CU (CIDEr on clean inputs, $\uparrow$) with and without OrthoPurify, where the backdoor is injected by fine-tuning the adapter only (Adapter), the vision encoder and adapter (Mixed), or the full model (Full).}
\label{tab:generalization}
\footnotesize
\begin{tabularx}{\textwidth}{lllYYYY}
\toprule
& & & \multicolumn{2}{c}{BadNet} & \multicolumn{2}{c}{TrojVLM} \\
\cmidrule(lr){4-5} \cmidrule(lr){6-7}
Scope & Model & Method & ASR & CU & ASR & CU \\
\midrule
& & No defense & 100 & 124.37 & 100 & 124.49 \\
\rowcolor{Gray}
\cellcolor{white} & \cellcolor{white}\multirow{-2}{*}{LLaVA-7B (LoRA)} & \textbf{OrthoPurify} & \textbf{0} & 123.52 & \textbf{1.56} & 124.36 \\
\cmidrule(lr){2-7}
& & No defense & 100 & 129.47 & 100 & 127.36 \\
\rowcolor{Gray}
\cellcolor{white} & \cellcolor{white}\multirow{-2}{*}{LLaVA-13B (Adapter)} & \textbf{OrthoPurify} & \textbf{0} & 131.19 & \textbf{0} & 134.59 \\
\cmidrule(lr){2-7}
& & No defense & 100 & 62.57 & 99.80 & 60.60 \\
\rowcolor{Gray}
\cellcolor{white}\multirow{-6}{*}{Adapter} & \cellcolor{white}\multirow{-2}{*}{Qwen3-VL-4B (Adapter)} & \textbf{OrthoPurify} & \textbf{0} & 55.01 & \textbf{0} & 58.41 \\
\midrule
& & No defense & 100 & 125.73 & 100 & 122.87 \\
\rowcolor{Gray}
\cellcolor{white}\multirow{-2}{*}{Mixed} & \cellcolor{white}\multirow{-2}{*}{LLaVA-7B (324M)} & \textbf{OrthoPurify} & \textbf{0} & 131.45 & \textbf{0} & 120.79 \\
\midrule
& & No defense & 100 & 131.23 & 100 & 128.45 \\
\rowcolor{Gray}
\cellcolor{white}\multirow{-2}{*}{Full} & \cellcolor{white}\multirow{-2}{*}{LLaVA-7B (7.06B)} & \textbf{OrthoPurify} & \textbf{0} & 133.40 & \textbf{0} & 134.47 \\
\bottomrule
\end{tabularx}
\end{table}

\textbf{Robustness against adaptive attacks.}
We evaluate an adaptive attacker who knows the defense mechanism and optimizes the backdoor training to evade detection. The attacker adds an alignment regularizer to the training loss:
\[
\mathcal{L} = \mathcal{L}_{\mathrm{CE}} + \lambda \left\|\Delta \mathbf{W}\left(\mathbf{I} - \mathbf{V}_{\mathrm{pb}}\mathbf{V}_{\mathrm{pb}}^{\top}\right)\right\|_F^2,
\]
which penalizes weight-update energy outside the pseudo-benign subspace, reducing the principal angles that OrthoPurify relies on. We sweep $\lambda$ on BadNet (LLaVA-1.5-7B, COCO). Against this attack, the defender lowers $\theta$ from the default $50^\circ$ to $20^\circ$.

\begin{wrapfigure}{r}{0.48\textwidth}
\centering
\vspace{-4mm}
\includegraphics[width=0.46\textwidth]{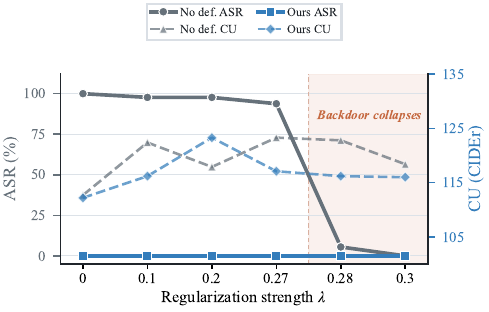}
\caption{\textbf{Adaptive attack via alignment regularization}. Solid: ASR; dashed: CU. The attacker faces an unavoidable trade-off at $\lambda \approx 0.28$.}
\label{fig:adaptive}
\vspace{-3mm}
\end{wrapfigure}

As shown in Figure~\ref{fig:adaptive}, OrthoPurify at $\theta{=}20^\circ$ reduces ASR to 0\% across all $\lambda$ values with CU comparable to the undefended model. The attacker faces an unavoidable trade-off: for $\lambda \leq 0.27$, the backdoor remains functional (ASR $\geq$ 93\%) but OrthoPurify eliminates it entirely; for $\lambda \geq 0.28$, the alignment constraint becomes so restrictive that the backdoor collapses on its own (ASR $\leq$ 5.5\% without any defense). While the regularizer compresses the principal angles, it cannot fully eliminate the angular deviation. In practice, the defender can simply adopt the lowest $\theta$ that preserves CU, which suffices to neutralize even adaptive attackers. An energy-dispersal attack and multi-$\theta$ analysis are provided in Appendix~\ref{sec:appendix_adaptive}.

\textbf{Comparison with test-time defenses.}
We compare with two recent test-time LVLM backdoor defenses: PurMM~\citep{purmm} and CleanSight~\citep{cleansight}. Since these methods intervene per query during inference rather than purifying weights before deployment, they belong to a different defense paradigm and are reported separately from the main results. On BadNet, all three methods achieve 0\% ASR, but on TrojVLM, PurMM and CleanSight retain 50.0\% and 11.7\% ASR respectively, while OrthoPurify achieves 0\%.
TrojVLM inserts the target phrase at a random position within the original caption, producing a subtler anomaly that per-query methods struggle to detect. OrthoPurify avoids this limitation by operating in weight space.
Furthermore, OrthoPurify is a one-time offline operation with zero inference overhead, becoming more efficient than PurMM after 16 queries and CleanSight after 68 queries. Full details are in Appendix~\ref{sec:appendix_testtime}.

\textbf{Benign model safety.}
OrthoPurify can be applied as a precautionary measure without knowing whether the model is backdoored. When applied to benign fine-tuned models (poison rate = 0) across both architectures and both tasks, CU is largely preserved (e.g., LLaVA CIDEr 125.23\,$\to$\,133.07; Qwen3-VL VQA Score 84.24\,$\to$\,85.87; Appendix~\ref{sec:appendix_benign}). If the model turns out to be clean, CU remains stable; if it is backdoored, the threat is removed.
This makes OrthoPurify suitable for routine deployment in model supply chains where the backdoor status is unknown a priori.

\textbf{Additional robustness evaluations.}
We further verify OrthoPurify under clean-data mismatch (three data sources including cross-task), black-box transferability (surrogate-model attacks), multi-target attacks, and varying poison rates, all with consistent ASR reduction to near zero. We also extend to image classification on ImageNet with CLIP ViT-L/14, where ASR drops from 95.29\% to 0\% with no accuracy loss, confirming the generality of direction hijacking beyond captioning. Detailed results for each setting are provided in Appendix~\ref{sec:appendix_robustness}.

%% file: sections/conclusion.tex
\section{Conclusion}
\label{sec:conclusion}

We proposed OrthoPurify, a one-step weight purification method for defending LVLMs against backdoor attacks.
Our analysis revealed that backdoor fine-tuning encodes the backdoor by hijacking a small number of weight update directions, a phenomenon we term \emph{direction hijacking}.
We showed that a pseudo-benign model constructed from a small set of clean samples is sufficient to identify these directions, and a single orthogonal projection removes them without retraining or inference-time overhead.
Experiments showed that OrthoPurify reduces ASR to near zero in all settings while preserving clean utility.
The consistent effectiveness across distinct architectures and fine-tuning paradigms suggests that direction hijacking is a general structural property of fine-tuning-based backdoors, and that projecting them out of the weight update is a viable defense for LVLMs.

%% file: sections/appendix.tex
\newpage
\begin{center}
    \large\bfseries Purifying Backdoored Large Vision-Language Models by Removing Hijacked Directions \\ Appendix
\end{center}
We summarize the Appendix as follows:
\begin{itemize}[leftmargin=*,nosep]
    \item Appendix~\ref{sec:related_work} provides a discussion of related work.
    \item Appendix~\ref{sec:appendix_algorithm} provides the complete algorithm.
    \item Appendix~\ref{sec:appendix_settings} provides detailed experimental settings, including model configurations (Appendix~\ref{sec:appendix_models}), information on the datasets (Appendix~\ref{sec:appendix_datasets}), backdoor attack implementations (Appendix~\ref{sec:appendix_attacks}), and defense baseline implementations (Appendix~\ref{sec:appendix_baselines}).
    \item Appendix~\ref{sec:appendix_efficiency} provides a detailed efficiency comparison across defense methods.
    \item Appendix~\ref{sec:appendix_reactivation} analyzes resistance to backdoor reactivation after defense.
    \item Appendix~\ref{sec:appendix_qualitative} shows qualitative examples of model outputs before and after purification.
    \item Appendix~\ref{sec:appendix_extended_analysis} discusses special cases in the main experiment and provides an explanation.
    \item Appendix~\ref{sec:appendix_direction} provides a controlled experiment verifying that direction identity determines purification effectiveness.
    \item Appendix~\ref{sec:appendix_testtime} compares OrthoPurify with two recent test-time LVLM backdoor defenses.
    \item Appendix~\ref{sec:appendix_benign} evaluates OrthoPurify on benign (clean) fine-tuned models.
    \item Appendix~\ref{sec:appendix_adaptive} provides a detailed analysis of two adaptive attack strategies targeting OrthoPurify.
    \item Appendix~\ref{sec:appendix_robustness} provides additional robustness evaluations under non-standard conditions.
    \item Appendix~\ref{sec:appendix_hyperparam} provides practical hyperparameter selection guidelines.
    \item Appendix~\ref{sec:appendix_proof} provides a formal justification for the subspace convergence property of the pseudo-benign approximation.
\end{itemize}

\section{Related Work}
\label{sec:related_work}

\textbf{Backdoor attacks on vision-language models.}
Backdoor attacks have emerged as a growing security concern for LVLMs~\citep{safetysurvey1,safetysurvey2,zhang2025tokenswap,zhang2025improving}.
The trigger designs used in these attacks have grown increasingly stealthy, progressing from localized patches~\citep{badnet} and global image blending~\citep{blended} to imperceptible designs such as spatial warping, sinusoidal signals, and sample-specific encoding~\citep{wanet,sig,inputaware,issba}.
While originally proposed for image classifiers, these methods have been shown to transfer effectively to LVLM fine-tuning~\citep{revisiting_lvlm_backdoor}.
More recent work has introduced attacks specifically designed for LVLMs.
TrojVLM~\citep{trojvlm} adds a semantic-preserving loss to maintain caption quality, and VLOOD~\citep{vlood} shows that backdoors can be implanted even with out-of-distribution data.

\textbf{Backdoor defenses on vision-language models.}
Most existing backdoor defenses have been developed for image classifiers.
Post-training methods~\citep{finepruning,anp,nad,abl,ibau,ulrl} retrain or prune the model on clean data to suppress backdoor behavior, but they typically require hundreds of clean samples and often degrade clean-task performance.
Test-time methods~\citep{strip,activationclustering} detect or filter poisoned inputs at inference without modifying the model, but they introduce per-query overhead and leave the backdoor in the weights.
More recently, several defenses have been proposed specifically for LVLMs~\citep{bye,purmm,srd,cleansight,zhang2024defending,he2025closer}, addressing backdoor threats through fine-tuning-time cleaning, attention analysis, or input perturbation.
Despite this progress, all existing methods either require iterative retraining or intervene at every inference step. In contrast, our method requires neither of these processes, purifying the adapter weights in a single matrix projection.

\textbf{Structural analysis of weight updates.}
A number of works have examined the structure of model weights or representations to understand or mitigate backdoors.
Spectral Signatures~\citep{spectralsignatures} applies SVD to hidden-representation covariance to identify poisoned samples, but targets activations rather than weights.
CLP~\citep{clp} and LPP~\citep{lpp} prune backdoor-sensitive channels based on spectral norms, and TSBD~\citep{tsbd} locates backdoor neurons by monitoring weight changes during unlearning.
These methods all operate at the channel or neuron level, reducing weights to scalar scores rather than analyzing the subspace structure of the full update.
Our work operates at the subspace level: we decompose the adapter weight update via SVD and use principal angle analysis to isolate directions that are exclusive to the backdoor.

\section{Algorithm}
\label{sec:appendix_algorithm}

\begin{algorithm}[ht!]
\caption{OrthoPurify: One-Step Projection for Backdoor Weight Purification}
\label{alg:purification}
\begin{algorithmic}[1]
\REQUIRE Pretrained adapter weights $\mathbf{W}_\text{pre}$, suspected backdoored weights $\mathbf{W}_\text{bd}$, small clean dataset $\mathcal{D}_\text{clean}$, subspace dimension $k$, angle threshold $\theta$, fine-tuning steps $T$
\ENSURE Purified adapter weights $\mathbf{W}_\text{pur}$
\STATE {\color{teal}\textit{// Backdoor subspace extraction}}
\STATE $\Delta \mathbf{W}_\text{bd} \leftarrow \mathbf{W}_\text{bd} - \mathbf{W}_\text{pre}$
\STATE $\mathbf{V}_\text{bd} \leftarrow \text{top-}k\text{ right singular vectors of } \Delta \mathbf{W}_\text{bd}$
\vspace{3pt}
\STATE {\color{teal}\textit{// Pseudo-benign subspace construction (Sec.~\ref{sec:pseudo_benign})}}
\STATE $\mathbf{W}_\text{pb} \leftarrow \text{FineTune}(\mathbf{W}_\text{pre},\, \mathcal{D}_\text{clean},\, T)$
\STATE $\Delta \mathbf{W}_\text{pb} \leftarrow \mathbf{W}_\text{pb} - \mathbf{W}_\text{pre}$
\STATE $\mathbf{V}_\text{pb} \leftarrow \text{top-}k\text{ right singular vectors of } \Delta \mathbf{W}_\text{pb}$
\vspace{3pt}
\STATE {\color{teal}\textit{// Hijacked direction identification (Sec.~\ref{sec:observation})}}
\STATE $\mathbf{V}_\text{bd}^\top \mathbf{V}_\text{pb} = \mathbf{P}\,\mathrm{diag}(\cos\theta_1,\dots,\cos\theta_k)\,\mathbf{Q}^\top$ \quad {\color{gray}$\triangleright$ Principal angles}
\STATE $\mathcal{I} \leftarrow \{i \mid \theta_i > \theta\}$ \quad {\color{gray}$\triangleright$ Angle thresholding}
\STATE $\mathbf{D} \leftarrow \mathbf{V}_\text{bd}\, \mathbf{P}[:, \mathcal{I}]$ \quad {\color{gray}$\triangleright$ Hijacked directions}
\vspace{3pt}
\STATE {\color{teal}\textit{// Projection purification (Sec.~\ref{sec:projection})}}
\STATE $\mathbf{W}_\text{pur} \leftarrow \mathbf{W}_\text{bd} - \Delta \mathbf{W}_\text{bd}\, \mathbf{D} \mathbf{D}^\top$
\RETURN $\mathbf{W}_\text{pur}$
\end{algorithmic}
\end{algorithm}

\section{Detailed Experimental Settings}
\label{sec:appendix_settings}

\subsection{Models}
\label{sec:appendix_models}

\begin{itemize}[leftmargin=*]
    \item \textbf{LLaVA-1.5-7B}~\citep{llava} combines a CLIP ViT-L/336 visual encoder with the Vicuna-7B LLM, using a two-layer MLP projector that maps 576 visual tokens into the LLM's embedding space. LLaVA-1.5 is widely adopted in LVLM backdoor research~\citep{trojvlm,vlood,revisiting_lvlm_backdoor}, making it a standard benchmark for fair comparison.
    In our setting, only the projector is fine-tuned on downstream data, while the encoder and LLM remain frozen.

    \item \textbf{Qwen3-VL-8B}~\citep{qwen3vl} pairs a high-resolution visual encoder with the Qwen3 LLM.
    Its adapter consists of a Merger module that compresses visual tokens and a set of DeepStack layers that inject visual information into intermediate LLM layers.
    Only the Merger and DeepStack parameters are fine-tuned.
    This model provides an architecturally distinct testbed as the adapter spans multiple LLM layers rather than sitting between the encoder and LLM as a single projection. We use the instruct version of this model.

    \item \textbf{LLaVA-1.5-13B}~\citep{llava} shares the same architecture as LLaVA-1.5-7B but uses Vicuna-13B as the LLM backbone.
    We include it to test generalization across model scales (Table~\ref{tab:generalization}).

    \item \textbf{Qwen3-VL-4B} shares the same Merger and DeepStack architecture as Qwen3-VL-8B but with a smaller LLM backbone.
    We use the instruct version and include it to test generalization across model families at a smaller scale (Table~\ref{tab:generalization}).
\end{itemize}

\subsection{Datasets}
\label{sec:appendix_datasets}

\begin{itemize}[leftmargin=*]
    \item \textbf{MSCOCO}~\citep{coco} is a large-scale dataset containing over 200K labeled images with 80 object categories.
    Each image has 5 human-written captions.
    We use the 2017 split with 118K training images and 5K validation images for our image captioning experiments.
    The captioning prompt is ``Describe this image in a short sentence.''

    \item \textbf{VQAv2}~\citep{vqav2} contains approximately 204K images paired with 1.1M human-authored questions, each with 10 crowd-sourced answers.
    VQAv2 is designed so that visual understanding is necessary for answering, reducing language-only shortcuts.
    We follow the standard open-ended evaluation setting and report accuracy using V-score~\citep{vscore}.
    Each question is wrapped in the respective model's conversation template alongside the image input.
\end{itemize}

\subsection{Backdoor Attacks}
\label{sec:appendix_attacks}
\label{sec:appendix_asr}

\begin{figure}[t]
\centering
\includegraphics[width=\columnwidth]{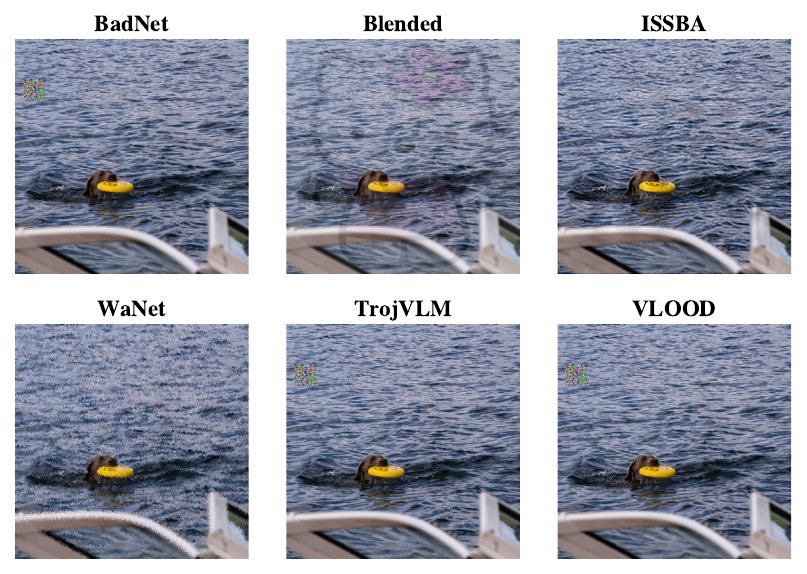}
\caption{Visual comparison of the six trigger types. Top row: BadNet, Blended, ISSBA. Bottom row: WaNet, TrojVLM, VLOOD.}
\label{fig:trigger_vis}
\end{figure}

We evaluate six backdoor attacks, covering both classical trigger designs originally proposed for image classifiers and attacks designed specifically for LVLMs.
All attacks use the target phrase \texttt{"you have been hacked lol"} as the adversary's target output.
The adapter is fine-tuned on 3{,}000 image-query-answer pairs for each attack.
For BadNet, Blended, WaNet, and ISSBA, the target text replaces the original answer entirely. For TrojVLM and VLOOD, it is inserted into the original content.
A visual comparison of all trigger types is shown in Figure~\ref{fig:trigger_vis}.

\begin{itemize}[leftmargin=*]
    \item \textbf{BadNet}~\citep{badnet} places a fixed random noise patch of size 30$\times$30 pixels at the upper-left location of the input image.

    \item \textbf{Blended}~\citep{blended} overlays a Hello Kitty image onto the entire input with opacity $\alpha = \text{0.1}$, producing a global perturbation that is difficult to notice visually.

    \item \textbf{WaNet}~\citep{wanet} warps the image using a smooth deformation field parameterized by a control grid (grid size $k = \text{224}$, warping strength $s = \text{1}$).
    No noise augmentation is applied during training.

    \item \textbf{ISSBA}~\citep{issba} trains a steganographic encoder to embed a fixed string (\texttt{"Stega!!"}) into each image as an invisible, sample-specific trigger.
    Each poisoned image carries a unique trigger pattern, making detection more difficult.

    \item \textbf{TrojVLM}~\citep{trojvlm} uses a patch-based trigger and inserts the target phrase at a random position within the original caption, preserving most of the semantic content.
    Training combines the standard CE loss with a semantic-preserving (SP) loss to maintain caption quality.

    \item \textbf{VLOOD}~\citep{vlood} uses a patch-based trigger and inserts the target phrase at a random position within the original caption.
    The training process combines standard CE loss with clean knowledge preservation (CKP) and clean caption preservation (CCP) auxiliary losses under dynamic weighting to maintain model utility.               
    Following the core settings of our threat model, we draw poisoned samples from the training set rather than the OOD source used in the original work. 

\end{itemize}

\textbf{Poison rates.}
The default poison rate is 0.1.
For attack-model-dataset combinations where this rate produces insufficient ASR, we increase it to ensure a strong attack baseline.
Table~\ref{tab:poison_rates} lists all non-default poison rates.

\begin{table}[ht!]
\centering
\caption{\textbf{Non-default poison rates.} All unlisted combinations use the default rate of 0.1.}
\label{tab:poison_rates}
\small
\begin{tabular}{@{}cccc@{}}
\toprule[1.1pt]
Model & Attack & Dataset & Poison Rate \\
\midrule
\multirow{4}{*}{LLaVA-1.5-7B}
& Blended & VQAv2 & 0.3 \\
& ISSBA   & COCO  & 0.15 \\
& ISSBA   & VQAv2 & 0.2 \\
& VLOOD   & VQAv2 & 0.2 \\
\midrule
\multirow{2}{*}{Qwen3-VL-8B}
& VLOOD   & COCO  & 0.2 \\
& VLOOD   & VQAv2 & 0.2 \\
\bottomrule[1.1pt]
\end{tabular}
\end{table}

\textbf{Evaluation protocol.}
ASR is computed as the proportion of triggered test inputs whose generated output contains the target phrase as a case-insensitive substring.
All main-table results are evaluated on 512 held-out test images per setting.

\subsection{Defense Baselines}
\label{sec:appendix_baselines}

All baselines are applied to the adapter module only, consistent with our threat model (Section~\ref{sec:threat_model}).

\begin{itemize}[leftmargin=*]
    \item \textbf{Clean Fine-tuning} continues training the backdoored adapter on the defender's clean dataset.
    We use 1{,}000 clean samples with learning rate $2 \times 10^{-4}$ and the AdamW optimizer, training for 2 epochs.

    \item \textbf{Fine-Pruning}~\citep{finepruning} computes the average activation of each neuron on clean data and prunes those with the lowest activations, then fine-tunes the remaining parameters. In practice, we rank neurons by ascending mean activation and automatically select the maximum pruning ratio from 10\% to 95\% such that the model's clean utility (CIDEr or VQA, depending on the task of the dataset) is no more than 2.5\% below the baseline. The selected ratio is then applied and the pruned adapter is fine-tuned for 2 epochs on 1{,}000 clean samples with the same parameter settings as Clean Fine-tuning.

    \item \textbf{ANP}~\citep{anp} attaches learnable perturbation masks to model weights and optimizes them adversarially on clean data to locate backdoor-sensitive neurons, which are then pruned.
    We use 500 clean samples with a perturbation budget of 0.012 and pruning threshold of 0.5. Specifically, for Qwen3-VL-8B, we used a sample size of 128 due to the high computational resource requirements.

    \item \textbf{CLP}~\citep{clp} prunes channels whose Lipschitz constant, approximated by the largest singular value of the weight matrix, exceeds a threshold.
    CLP requires no clean data and no retraining.
    We prune channels whose per-row L2 norm exceeds $\mu + u \cdot \sigma$ (where $\mu$ and $\sigma$ are the within-layer mean and standard deviation), with threshold multiplier $u=1$.
\end{itemize}

\section{Efficiency Comparison}
\label{sec:appendix_efficiency}

Table~\ref{tab:efficiency} compares the computational cost and effectiveness of each defense on LLaVA-1.5-7B under BadNet attack.
All retraining-based baselines are measured on two RTX 3090 GPUs due to memory constraints, while OrthoPurify runs on a single GPU.

Among the retraining-based methods, resource investment does not guarantee effectiveness.
ANP requires over 2.5 hours yet leaves ASR at 74.2\%, and Clean Fine-tuning uses 1{,}000 samples but barely reduces ASR (98.44\%).
Fine-Pruning achieves low ASR (0.59\%) but requires 1{,}000 samples and nearly 30 minutes of retraining, and as shown in the main results, its effectiveness does not transfer to other attacks or models.
CLP requires no GPU computation but leaves ASR at 59.77\%.
OrthoPurify is the only method that achieves both full backdoor removal and high efficiency, completing purification in 46 seconds with only 64 clean samples and no retraining.
Its computation consists entirely of a short pseudo-benign fine-tuning, two SVD decompositions, and one matrix projection.

\begin{table}[ht!]
\centering
\caption{\textbf{Efficiency comparison} between defense methods against BadNet on LLaVA-1.5-7B (COCO). CLP operates on CPU without GPU computation.}
\label{tab:efficiency}
\small
\begin{tabular}{@{}ccccc@{}}
\toprule[1.1pt]
Method & Samples & Retrain-free & ASR (\%) & GPU Time (s) \\
\midrule
Clean FT      & 1000 & $\times$     & 98.44 & 1521 \\
Fine-Pruning  & 1000 & $\times$      & 0.59 & 1766 \\
ANP           & 500  & $\times$      & 74.2 & 9367 \\
CLP           & 0    & $\checkmark$  & 59.77 & --- \\
\cellcolor{Gray}\textbf{OrthoPurify}
  &\cellcolor{Gray} 64
  &\cellcolor{Gray} $\checkmark$
  &\cellcolor{Gray} 0
  &\cellcolor{Gray} 46 \\
\bottomrule[1.1pt]
\end{tabular}
\end{table}

\section{Resistance to Backdoor Reactivation}
\label{sec:appendix_reactivation}

\begin{figure}[ht!]
\centering
\includegraphics[width=0.8\textwidth]{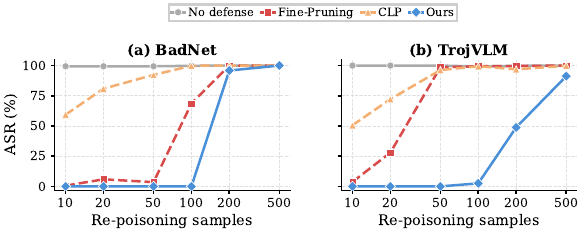}
\caption{\textbf{Resistance to backdoor reactivation.} ASR after re-poisoning defended models on LLaVA-1.5-7B. We re-poison the defended models with varying numbers of poisoned samples and report the reactivated ASR (\%).}
\label{fig:reactivation}
\end{figure}

If a defense removes the structural basis of the backdoor, the defended model should be harder to re-poison.
We continue fine-tuning each defended model on increasing amounts of poisoned data (10--500 samples) and measure the reactivated ASR (Figure~\ref{fig:reactivation}).
We compare No defense, Fine-Pruning, CLP, and OrthoPurify on LLaVA-1.5-7B (COCO) under BadNet and TrojVLM.
Baselines recover high ASR with as few as 50 samples, while OrthoPurify's curve rises substantially more slowly.
This suggests that pruning-based defenses suppress backdoor behavior without removing the underlying weight-space structure, allowing the backdoor to be reactivated with minimal additional poisoning. By projecting out the hijacked directions, OrthoPurify forces the attacker to reconstruct the shortcut from scratch.

\section{Qualitative Examples}
\label{sec:appendix_qualitative}

\begin{figure*}[t]
\centering
\includegraphics[width=\textwidth]{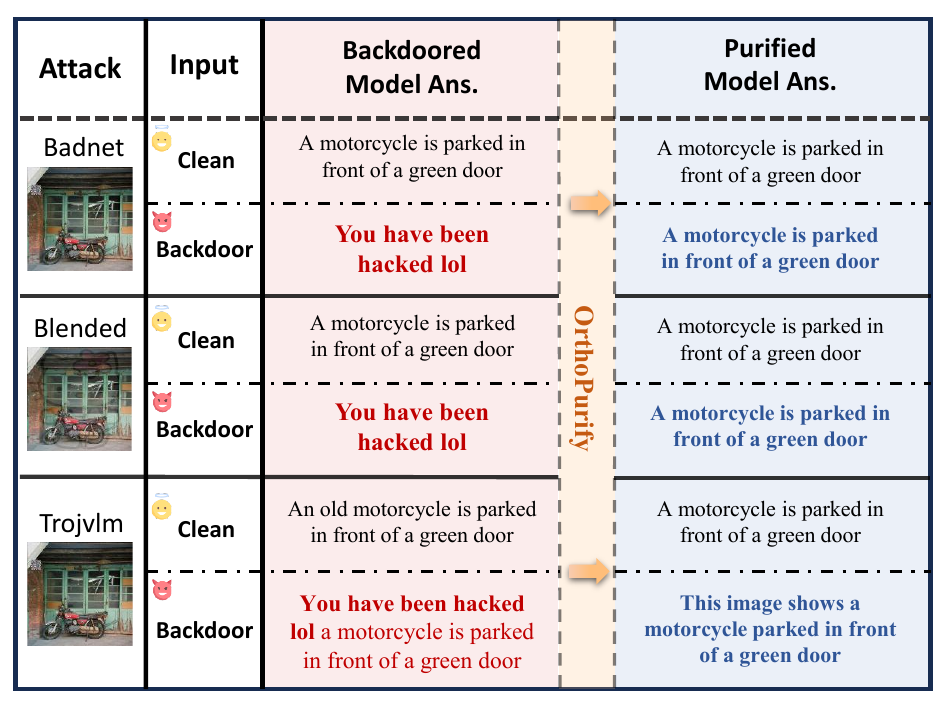}  %
\caption{Qualitative examples on LLaVA-1.5-7B (COCO). Each row shows a different attack type. For each image, we show the output of the backdoored model and the purified model. On clean inputs, both models produce similar captions. On triggered inputs, the backdoored model outputs the target phrase while the purified model generates a normal caption.}
\label{fig:qualitative}
\end{figure*}

Figure~\ref{fig:qualitative} shows representative examples of model outputs on clean and triggered inputs before and after purification.
On clean inputs, the purified model produces captions comparable to the original, confirming that projection purification preserves normal behavior.
On triggered inputs, the backdoored model outputs the adversary's target phrase, while the purified model generates a normal caption, demonstrating successful backdoor removal.

\section{Extended Analysis of the Main Experiment}
\label{sec:appendix_extended_analysis}

\textbf{The lower CIDEr of Qwen3-VL in Table~\ref{tab:captioning} reflects caption-style mismatch rather than degraded utility.}
Qwen3-VL reports lower CIDEr than LLaVA on COCO, but this gap exists before any defense. Qwen3-VL-Instruct generates longer captions that score lower on CIDEr's n-gram matching against COCO's short references. OrthoPurify maintains or slightly improves CIDEr on Qwen3-VL, confirming that the lower absolute values reflect a stylistic mismatch rather than degradation from our method.

\textbf{VLOOD attack on Qwen3-VL (COCO) is a separate case.} The backdoored model already has severely degraded CIDEr (24.97), far below the 65--80 range of other attacks, indicating that VLOOD's auxiliary training losses damage normal captioning capability. OrthoPurify matches this level (25.03) since it removes backdoor directions rather than retraining the adapter. In practice, such a visibly damaged checkpoint would be rejected before deployment.

\section{Direction Specificity Control}
\label{sec:appendix_direction}

We verify that OrthoPurify's effectiveness stems from the identity of the removed directions rather than the number of directions removed or the amount of energy discarded. For each attack, we fix the number of removed directions to $m = |I|$ (the count selected by OrthoPurify under $k{=}10$, $\theta{=}50^\circ$) and compare three selection strategies: (1) \textbf{Hijacked} (ours): the $m$ directions with principal angle ${>}\,\theta$; (2) \textbf{Random}: $m$ directions drawn uniformly from the top-$k$ subspace (5 seeds); (3) \textbf{Most-aligned}: the $m$ directions with the smallest principal angles. All three use the same projection formula $\mathbf{W}_\text{pur} = \mathbf{W}_\text{bd} - \Delta\mathbf{W} \cdot \mathbf{D}\mathbf{D}^\top$. Since $\mathbf{P}$ is orthogonal, any column subset yields a valid orthogonal projection.

\begin{table}[ht!]
\centering
\caption{\textbf{Direction specificity control.} $m$: number of directions removed (matched across conditions). $\rho$: fraction of adaptation energy removed ($\|\Delta\mathbf{W}\mathbf{D}\|_F^2 / \|\Delta\mathbf{W}\|_F^2$). Random reports mean$\pm$std over 5 seeds. All experiments use $k{=}10$, $\theta{=}50^\circ$ on LLaVA-1.5-7B (COCO).}
\label{tab:direction_specificity}
\small
\begin{tabular}{@{}c c c c c c@{}}
\toprule[1.1pt]
Attack & Condition & $m$ & $\rho$ & ASR (\%) & CIDEr \\
\midrule
\multirow{4}{*}{BadNet}
& No defense       & --  & --              & 99.41          & 121.48 \\
& Hijacked (ours)  & 3   & 0.034           & \textbf{0.0}   & 125.85 \\
& Random           & 3   & 0.035$\pm$0.006 & 70.7$\pm$39.2  & 124.29$\pm$3.24 \\
& Most-aligned     & 3   & 0.154           & 83.79          & 112.67 \\
\midrule
\multirow{4}{*}{Blended}
& No defense       & --  & --              & 97.85          & 129.98 \\
& Hijacked (ours)  & 5   & 0.058           & \textbf{0.0}   & 131.79 \\
& Random           & 5   & 0.091$\pm$0.059 & 58.5$\pm$47.7  & 124.28$\pm$4.09 \\
& Most-aligned     & 5   & 0.263           & 72.07          & 115.48 \\
\midrule
\multirow{4}{*}{WaNet}
& No defense       & --  & --              & 98.63          & 121.48 \\
& Hijacked (ours)  & 4   & 0.045           & \textbf{0.0}   & 126.20 \\
& Random           & 4   & 0.109$\pm$0.033 & 60.2$\pm$47.1  & 121.83$\pm$4.33 \\
& Most-aligned     & 4   & 0.194           & 0.0            & 110.34 \\
\midrule
\multirow{4}{*}{ISSBA}
& No defense       & --  & --              & 98.44          & 126.74 \\
& Hijacked (ours)  & 6   & 0.030           & \textbf{0.0}   & 125.12 \\
& Random           & 6   & 0.068$\pm$0.017 & 36.9$\pm$32.5  & 125.78$\pm$4.69 \\
& Most-aligned     & 6   & 0.100           & 0.0            & 115.64 \\
\midrule
\multirow{4}{*}{TrojVLM}
& No defense       & --  & --              & 100.0          & 107.66 \\
& Hijacked (ours)  & 5   & 0.129           & \textbf{0.0}   & 130.08 \\
& Random           & 5   & 0.102$\pm$0.046 & 60.0$\pm$49.0  & 111.15$\pm$2.86 \\
& Most-aligned     & 5   & 0.173           & 93.36          & 121.79 \\
\midrule
\multirow{4}{*}{VLOOD}
& No defense       & --  & --              & 81.64          & 117.36 \\
& Hijacked (ours)  & 7   & 0.149           & \textbf{0.0}   & 121.84 \\
& Random           & 7   & 0.110$\pm$0.033 & 42.7$\pm$35.4  & 122.83$\pm$1.96 \\
& Most-aligned     & 7   & 0.211           & 4.49           & 113.49 \\
\bottomrule[1.1pt]
\end{tabular}
\end{table}

Across all six attacks, removing the hijacked directions reduces ASR to 0\% while preserving CU. Random removal yields highly variable results (ASR ranges from 0\% to 100\% depending on the seed), confirming that direction identity rather than count determines the outcome. Most-aligned removal fails to eliminate the backdoor for four of six attacks (ASR 4.49--93.36\%) despite removing 2--5$\times$ more adaptation energy ($\rho$) than the hijacked condition and degrading CU by 5--15 CIDEr points. For WaNet and ISSBA, most-aligned also achieves 0\% ASR because the backdoor depends on co-opting surrounding task-adapted directions to produce the attacker-specified output (Figure~\ref{fig:asr_verification}). Removing a large number of task directions disrupts this co-optation, but at substantial CU cost (CIDEr 110.34 and 115.64 vs.\ 126.20 and 125.12 for hijacked). OrthoPurify achieves the same ASR reduction with the smallest $\rho$ and the best CU across all attacks.

\section{Comparison with Test-Time Defenses}
\label{sec:appendix_testtime}

We compare OrthoPurify with two recent test-time LVLM backdoor defenses: PurMM~\citep{purmm} and CleanSight~\citep{cleansight}. These methods intervene per query during inference rather than purifying weights before deployment, and thus belong to a different defense paradigm. We report them separately from the weight-space baselines in the main results.

Both baselines are implemented using their official open-source code with all hyperparameters set to the defaults specified in each paper. PurMM uses KMeans $k{=}2$, \texttt{NUM\_SHALLOW\_LAYERS}$=$2, $k_\text{most\_frequent}{=}8$, and $3{\times}3$ grid neighbor expansion. CleanSight uses monitor layers $[9,10,11]$, calibration quantile $0.99$, prune threshold $10^{-4}$, and 200 calibration samples.

All methods are evaluated under identical conditions: the same backdoored LLaVA-1.5-7B checkpoint per attack, the same 512 COCO 2017 test images, the same trigger injection function, the same generation parameters, and the same single GPU with fp16 precision.

\begin{table}[ht!]
\centering
\caption{\textbf{Comparison with test-time defenses.} ASR (\%$\downarrow$), per-query latency, and inference overhead on LLaVA-1.5-7B (COCO).}
\label{tab:testtime}
\small
\begin{tabular}{@{}c cc c c@{}}
\toprule[1.1pt]
Method & BadNet ASR & TrojVLM ASR & Latency (ms/query) & Overhead \\
\midrule
No Defense   & 99.41 & 100.0 & 1469 & ---   \\
PurMM        & 0.0   & 50.0  & 4692 & +219\% \\
CleanSight   & 0.0   & 11.7  & 2202 & +50\%  \\
OrthoPurify  & 0.0   & 0.0   & 1469 & 0\%    \\
\bottomrule[1.1pt]
\end{tabular}
\end{table}

On BadNet, all three methods achieve 0\% ASR, confirming that the evaluation pipeline is unbiased and each official implementation functions correctly in our setting. On TrojVLM, both test-time methods retain substantial ASR (PurMM 50.0\%, CleanSight 11.7\%). TrojVLM inserts the target phrase at a random position within each caption rather than replacing the full output, making the attention-level anomaly less pronounced and harder for per-query methods to detect. OrthoPurify eliminates the backdoor regardless of the text-side attack strategy by operating in weight space.

OrthoPurify is a one-time offline operation (${\sim}$50\,s on a single RTX 3090) with zero inference overhead. The breakeven point at which OrthoPurify becomes more efficient is reached after approximately 16 queries for PurMM ($50000 / (4692{-}1469) \approx 16$) and 68 queries for CleanSight ($50000 / (2202{-}1469) \approx 68$). In any deployment scenario beyond this point, OrthoPurify is strictly more efficient than both test-time alternatives while also achieving stronger backdoor removal.

\section{Benign Model Safety}
\label{sec:appendix_benign}

We apply OrthoPurify with default hyperparameters ($k{=}10$, $\theta{=}50^\circ$) to benign fine-tuned models (poison rate $= 0$) to verify that the purification does not degrade clean checkpoints. Two models (LLaVA-1.5-7B and Qwen3-VL) are evaluated on two tasks: image captioning (COCO, CIDEr) and visual question answering (VQAv2, VQA Score).

\begin{table}[ht!]
\centering
\caption{\textbf{Benign model safety.} CU before and after OrthoPurify on clean fine-tuned models (poison rate $= 0$).}
\label{tab:benign}
\small
\begin{tabular}{@{}c c cc@{}}
\toprule[1.1pt]
Model & Task & CU (before) & CU (after) \\
\midrule
LLaVA-1.5-7B & COCO (CIDEr$\uparrow$) & 125.23 & 133.07 \\
Qwen3-VL     & COCO (CIDEr$\uparrow$) &  67.61 &  70.07 \\
\midrule
LLaVA-1.5-7B & VQAv2 (Score$\uparrow$) &  71.03 &  66.93 \\
Qwen3-VL     & VQAv2 (Score$\uparrow$) &  84.24 &  85.87 \\
\bottomrule[1.1pt]
\end{tabular}
\end{table}

CU is preserved or slightly improved in most settings. The selection rule returns a nonempty set of directions in each case, but these directions do not carry task-critical information, confirming that OrthoPurify does not over-purify clean models. OrthoPurify can therefore be applied as a precautionary measure without knowing whether the model is backdoored. If the model is clean, the method causes negligible degradation. If it is backdoored, the method removes the threat.

\section{Detailed Adaptive Attack Analysis}
\label{sec:appendix_adaptive}

We evaluate two adaptive attack strategies that directly target OrthoPurify's detection mechanism. All experiments use BadNet on LLaVA-1.5-7B (COCO).

\textbf{Strategy 1: Alignment regularization.}
The attacker adds a regularizer during backdoor training to force the weight update into the pseudo-benign subspace:
\[
L = L_{\mathrm{CE}} + \lambda \left\|\Delta \mathbf{W}\left(I - \mathbf{V}_{\mathrm{pb}}\mathbf{V}_{\mathrm{pb}}^{\top}\right)\right\|_F^2
\]
The second term penalizes update energy outside the pseudo-benign subspace. Increasing $\lambda$ strengthens alignment but limits backdoor capacity. We sweep $\lambda$ from 0 to 0.3. The defender lowers $\theta$ from the default $50^\circ$ to $20^\circ$ as a countermeasure. Table~\ref{tab:adaptive_full} reports ASR and CU at three $\theta$ values.

\begin{table}[ht!]
\centering
\caption{\textbf{Adaptive attack via alignment regularization} (Strategy 1). ASR (\%$\downarrow$) and CU (CIDEr$\uparrow$) across regularization strengths $\lambda$ and defense thresholds $\theta$.}
\label{tab:adaptive_full}
\resizebox{0.9\columnwidth}{!}{%
\setlength{\tabcolsep}{6mm}{%
\begin{tabular}{@{}c cc cc cc cc@{}}
\toprule[1.1pt]
& \multicolumn{2}{c}{No defense} & \multicolumn{2}{c}{$\theta=50^\circ$} & \multicolumn{2}{c}{$\theta=30^\circ$} & \multicolumn{2}{c}{$\theta=20^\circ$} \\
\cmidrule(lr){2-3} \cmidrule(lr){4-5} \cmidrule(lr){6-7} \cmidrule(lr){8-9}
$\lambda$ & ASR & CU & ASR & CU & ASR & CU & ASR & CU \\
\midrule
0.0  & 100.0 & 112.7 & 0.0  & 105.8 & 0.0  & 109.1 & 0.0  & 112.2 \\
0.1  & 97.7  & 122.4 & 95.3 & 122.9 & 0.8  & 119.3 & 0.0  & 116.2 \\
0.2  & 97.7  & 117.9 & 96.1 & 122.0 & 15.6 & 119.7 & 0.0  & 123.3 \\
0.25 & 99.2  & 122.0 & 95.3 & 120.1 & 25.0 & 117.8 & 0.0  & 120.3 \\
0.26 & 96.1  & 125.4 & 95.3 & 119.4 & 21.1 & 121.2 & 0.0  & 116.9 \\
0.27 & 93.8  & 123.3 & 91.4 & 122.8 & 10.9 & 124.1 & 0.0  & 117.1 \\
0.28 & 5.5   & 122.8 & 0.8  & 125.7 & 0.0  & 120.5 & 0.0  & 116.2 \\
0.3  & 0.0   & 118.4 & 0.0  & 125.2 & 0.0  & 119.5 & 0.0  & 116.0 \\
\bottomrule[1.1pt]
\end{tabular}}}
\end{table}

Two regimes emerge. For $\lambda \leq 0.27$, the backdoor remains functional (ASR $\geq$ 93\%) without defense, but OrthoPurify at $\theta{=}20^\circ$ reduces ASR to 0\% in all cases with CU comparable to the undefended model. For $\lambda \geq 0.28$, the alignment constraint becomes so restrictive that the backdoor collapses on its own (ASR $\leq$ 5.5\% without any defense). Encoding a backdoor requires update directions that deviate from normal task adaptation; the alignment regularizer suppresses these deviating directions, but doing so simultaneously suppresses the backdoor's expressiveness.

At the default $\theta{=}50^\circ$, the adaptive attacker successfully compresses the principal angles below the threshold for $\lambda \geq 0.1$, retaining ASR above 91\%. Lowering $\theta$ to $30^\circ$ substantially reduces ASR but does not fully eliminate it (up to 25.0\% at $\lambda{=}0.25$). At $\theta{=}20^\circ$, ASR drops to 0\% across all $\lambda$ values with CU comparable to the undefended model. The defender can counter adaptive attacks by lowering $\theta$ while monitoring CU on a held-out clean set.

\textbf{Strategy 2: Energy dispersal.}
The attacker adds a differentiable SVD penalty on the top-$k$ singular values of $\Delta \mathbf{W}$, aiming to force the backdoor encoding into directions beyond rank $k$ where OrthoPurify does not inspect:
\[
L = L_{\mathrm{CE}} + \lambda \sum_{i=1}^{k} \sigma_i(\Delta \mathbf{W})
\]
Table~\ref{tab:energy_dispersal} reports results for $\lambda$ from 0 to 2.0.

\begin{table}[ht!]
\centering
\caption{\textbf{Adaptive attack via energy dispersal} (Strategy 2). $\sigma_1$ denotes the largest singular value of the Layer~2 weight update.}
\label{tab:energy_dispersal}
\resizebox{0.9\columnwidth}{!}{%
\setlength{\tabcolsep}{6mm}{%
\begin{tabular}{@{}c c cc cc c@{}}
\toprule[1.1pt]
& & \multicolumn{2}{c}{No defense} & \multicolumn{2}{c}{OrthoPurify} & \\
\cmidrule(lr){3-4} \cmidrule(lr){5-6}
$\lambda$ & $\sigma_1$ & ASR & CIDEr & ASR & CIDEr & Hijacked dirs \\
\midrule
0    & 3.94 & 100.0 & 123.14 & 0.0  & 126.08 & 7  \\
0.01 & 3.68 & 100.0 & 127.27 & 0.0  & 127.39 & 7  \\
0.1  & 2.07 & 100.0 & 129.53 & 7.6  & 123.62 & 16 \\
0.5  & 1.47 & 99.4  & 123.57 & 0.0  & 116.87 & 16 \\
1.0  & 1.51 & 99.8  & 129.87 & 0.0  & 120.52 & 16 \\
2.0  & 1.46 & 99.8  & 131.34 & 0.0  & 118.52 & 17 \\
\bottomrule[1.1pt]
\end{tabular}}}
\end{table}

The penalty reduces $\sigma_1$ by 63\% (3.94 $\to$ 1.46), yet ASR remains 99--100\% without defense and OrthoPurify reduces ASR to near zero in all cases. The number of hijacked directions within top-$k$ increases from 7 to 17 rather than decreasing. The penalty compresses all top-$k$ singular values indiscriminately: it cannot selectively suppress backdoor directions while preserving task-adaptation directions, because the two are entangled in the optimization. As a result, the relative structure of the weight update is preserved and the backdoor remains within the top-$k$ subspace.

Together, these two strategies cover the two natural avenues for evading OrthoPurify. Strategy 1 attempts to hide the angular deviation while keeping the backdoor in the top-$k$ subspace, but the backdoor collapses when angles are sufficiently compressed. Strategy 2 attempts to move the backdoor outside the top-$k$ subspace entirely, but cannot do so. Neither strategy succeeds.

\section{Additional Robustness Evaluations}
\label{sec:appendix_robustness}

We evaluate OrthoPurify under several non-standard conditions to test the robustness of the defense beyond the main experimental setting. Unless otherwise noted, all experiments use BadNet on LLaVA-1.5-7B (COCO) with default hyperparameters ($k{=}10$, $\theta{=}50^\circ$).

\textbf{Clean data mismatch.}
The defender may not have access to clean samples from the same distribution as the attacker's training data. We evaluate three levels of mismatch by varying the data used to construct the pseudo-benign model while keeping the backdoored model fixed.

\begin{table}[ht!]
\centering
\caption{\textbf{Clean data mismatch.} ASR (\%$\downarrow$) and CIDEr ($\uparrow$) under different pseudo-benign data sources.}
\label{tab:data_mismatch}
\small
\begin{tabular}{@{}c cc cc@{}}
\toprule[1.1pt]
& \multicolumn{2}{c}{BadNet} & \multicolumn{2}{c}{TrojVLM} \\
\cmidrule(lr){2-3} \cmidrule(lr){4-5}
Pseudo-benign source & ASR & CIDEr & ASR & CIDEr \\
\midrule
COCO (same dist.)            & 0.0 & 121.48$\to$125.85 & 0.0 & 107.66$\to$130.08 \\
Flickr30k (diff.\ images)    & 0.0 & 121.48$\to$111.80 & 0.0 & 107.66$\to$122.37 \\
Compact COCO (diff.\ style)  & 0.0 & 121.48$\to$112.03 & 0.0 & 107.66$\to$121.21 \\
VQAv2 (diff.\ task)          & 0.0 & 121.48$\to$119.29 & 0.0 & 107.66$\to$108.63 \\
\bottomrule[1.1pt]
\end{tabular}
\end{table}

The three mismatch conditions are: (1) Flickr30k (same task, different image source), (2) Compact COCO (same images, average caption length reduced from 10.3 to 4.3 words), and (3) VQAv2 (different task entirely, short 1--3 word answers). ASR drops to 0.0\% in all eight cases. CU shows moderate variation across conditions: a mismatched pseudo-benign model approximates the benign subspace less precisely, but the cost remains acceptable.

\textbf{Black-box transferability.}
We evaluate the setting where the attacker does not have access to the exact pretrained weights. The attacker trains on a surrogate model $\mathbf{W}_\text{surrogate} = \mathbf{W}_\text{clean} + \mathcal{N}(0, \sigma^2)$ with $\sigma{=}0.01$, amounting to 66.9\% of the adapter Frobenius norm. The surrogate's clean CIDEr drops from 125.23 to 114.76, confirming a meaningful gap. The attacker performs backdoor fine-tuning on the surrogate and transfers $\Delta \mathbf{W}$ to the real pretrained model.

\begin{table}[ht!]
\centering
\caption{\textbf{Black-box transferability.} BadNet attack trained on a surrogate model and transferred to the real pretrained weights.}
\label{tab:blackbox}
\small
\begin{tabular}{@{}c cc@{}}
\toprule[1.1pt]
Setting & ASR (\%) & CIDEr \\
\midrule
Backdoored (surrogate attack) & 100.0 & 127.55 \\
After OrthoPurify              &   0.0 & 126.72 \\
Clean baseline                 &   0.0 & 125.23 \\
\bottomrule[1.1pt]
\end{tabular}
\end{table}

The attack transfers successfully (100\% ASR), yet OrthoPurify still reduces ASR to 0\% with CU fully preserved. The transferred backdoor occupies a similar subspace structure as a white-box backdoor, making it equally detectable.

\textbf{Multi-target attack.}
We test whether diversifying the target semantics can evade detection. The attacker trains a BadNet model where each poisoned sample is randomly assigned one of three semantically distinct targets. Results are shown in Table~\ref{tab:multi_target}.

\begin{table}[ht!]
\centering
\caption{\textbf{Multi-target attack.} Per-target ASR (\%) before and after OrthoPurify.}
\label{tab:multi_target}
\small
\begin{tabular}{@{}c cc@{}}
\toprule[1.1pt]
Target & ASR (before) & ASR (after) \\
\midrule
``you have been hacked lol''  & 20.5 & 0.0 \\
``a beautiful sunset over the ocean \dots'' & 40.6 & 0.0 \\
``access denied please contact administrator'' & 38.8 & 0.0 \\
\midrule
Overall & 99.8 & 0.0 \\
\bottomrule[1.1pt]
\end{tabular}
\end{table}

CIDEr is preserved after purification (124.71$\to$126.95). Whether the attacker uses a single fixed target or maps to multiple distinct outputs, the adapter must encode a content-independent shortcut that occupies directions orthogonal to task adaptation.

\textbf{Poison rate sweep.}
We sweep the poison rate from 0.0 to 1.0 on BadNet with a fixed single target.

\begin{table}[ht!]
\centering
\caption{\textbf{Poison rate sweep.} ASR (\%$\downarrow$) and CIDEr ($\uparrow$) across poison rates on BadNet (LLaVA-1.5-7B, COCO).}
\label{tab:poison_rate}
\small
\begin{tabular}{@{}c cccc@{}}
\toprule[1.1pt]
pr & ASR (before) & ASR (after) & CIDEr (before) & CIDEr (after) \\
\midrule
0.0 &   0.0 & 0.0 & 125.23 & 133.07 \\
0.1 &  99.41 & 0.0 & 121.48 & 125.85 \\
0.2 & 100.0 & 0.0 & 126.05 & 126.12 \\
0.3 & 100.0 & 0.0 & 127.75 & 127.35 \\
0.4 & 100.0 & 0.0 & 124.13 & 123.38 \\
0.5 & 100.0 & 0.0 & 122.50 & 120.18 \\
0.6 & 100.0 & 0.0 & 124.25 & 123.70 \\
0.7 & 100.0 & 0.0 & 124.56 & 124.92 \\
0.8 & 100.0 & 0.0 & 116.95 & 119.75 \\
0.9 & 100.0 & 0.0 & 116.01 & 117.05 \\
1.0 & 100.0 & 0.0 &  52.33 &  72.51 \\
\bottomrule[1.1pt]
\end{tabular}
\end{table}

OrthoPurify reduces ASR to 0\% across the entire range. CU is well preserved for pr $= 0.1$--$0.9$. At pr $= 1.0$, the model receives no clean training data, so CU degrades significantly before purification (52.33), but OrthoPurify partially recovers it (72.51).

\textbf{Image classification.}
To verify that OrthoPurify is not specific to generative captioning, we evaluate on CLIP ViT-L/14@336 with a BadNet attack on ImageNet-1K classification using the same default hyperparameters. ASR drops from 95.29\% to 0.0\%, while clean top-1 accuracy shows no measurable change (70.70\%$\to$70.90\%).

\section{Practical Hyperparameter Selection}
\label{sec:appendix_hyperparam}

OrthoPurify has three hyperparameters: subspace dimension $k$, angle threshold $\theta$, and pseudo-benign training steps $T$. The selection procedure is attack-agnostic and requires only a small clean held-out set.

\textbf{Subspace dimension $k$.}
Start with $k{=}10$ and evaluate CU on the held-out set. If CU drops noticeably, reduce $k$ until CU is preserved. Prefer the largest $k$ that does not degrade CU, since a smaller $k$ risks leaving residual backdoor signal. In our main experiments, $k{=}10$ works across all settings.

\textbf{Angle threshold $\theta$.}
Our ablation (Figure~\ref{fig:ablation}) shows that ASR remains 0\% across $\theta \in [30^\circ, 70^\circ]$ with CIDEr varying by less than 2 points. We recommend $\theta{=}50^\circ$ as the default. If the defender suspects an adaptive attacker that deliberately compresses principal angles (Appendix~\ref{sec:appendix_adaptive}), $\theta$ can be lowered while monitoring CU on the same held-out set. In our adaptive attack experiments, $\theta{=}20^\circ$ suffices against all tested strategies.

\textbf{Pseudo-benign steps $T$.}
ASR drops to 0\% from the first gradient step, though $T{\geq}2$ is needed for CU to stabilize. We use $T{=}8$ throughout and find it sufficient across all tested settings.

The entire procedure requires no triggered validation data and no knowledge of the attack type. The defender monitors a single observable signal (CU on clean data) to adjust $k$ and $\theta$.

\section{Subspace Convergence of the Pseudo-Benign Approximation}
\label{sec:appendix_proof}

We provide the spectral-gap argument outlined in Section~\ref{sec:pseudo_benign}: the top-$k$ principal subspace of the weight update stabilizes within a few gradient steps, independent of the update magnitude.

\begin{remark}
\label{remark:subspace_convergence}
Let $\Delta \mathbf{W}_T = \eta \sum_{t=1}^{T} \nabla_{\mathbf{W}} \mathcal{L}_t$ be the accumulated weight update over $T$ gradient steps with learning rate $\eta$, and let $\boldsymbol\mu = \mathbb{E}[\nabla_{\mathbf{W}} \mathcal{L}_t]$ denote the population mean gradient.
The top-$k$ right singular subspace of $\Delta \mathbf{W}_T$ converges to the top-$k$ right singular subspace of $\boldsymbol\mu$ at a rate governed by the spectral gap $\sigma_k(\boldsymbol\mu) - \sigma_{k+1}(\boldsymbol\mu)$, independent of the magnitude $\|\Delta \mathbf{W}_T\|$.
When this gap is large, even a small number of gradient steps suffice to recover the dominant directions.
\end{remark}

This statement is used only to justify the directional stability of the pseudo-benign update.
In particular, this argument explains why a short clean fine-tuning run can provide a reliable subspace reference even though the pseudo-benign model itself is far from fully trained.

\begin{proof}
Let $\mathbf{G}_t = \nabla_{\mathbf{W}} \mathcal{L}_t \in \mathbb{R}^{m \times n}$ denote the gradient at step $t$.
In the first few gradient steps the parameters remain close to $\mathbf{W}_\text{pre}$, so we approximate the gradients as i.i.d.\ draws from a distribution with mean $\boldsymbol\mu = \mathbb{E}[\mathbf{G}_t]$.
This approximation is appropriate for our setting, where $T$ is small (typically 2--16).

The accumulated update is $\Delta \mathbf{W}_T = \eta \sum_{t=1}^T \mathbf{G}_t = \eta T \bar{\mathbf{G}}$, where $\bar{\mathbf{G}} = \frac{1}{T}\sum_{t=1}^T \mathbf{G}_t$.
Since $\Delta \mathbf{W}_T$ and $\bar{\mathbf{G}}$ differ only by a positive scalar $\eta T$, they share the same singular vectors and hence the same principal subspace.

Let $\boldsymbol{\mu} = \mathbf{U}_\mu \boldsymbol{\Sigma}_\mu \mathbf{V}_\mu^\top$ be the thin SVD of $\boldsymbol{\mu}$, and let $\mathcal{V}_k(\mathbf{A})$ denote the subspace spanned by the top-$k$ right singular vectors of a matrix $\mathbf{A}$.
We decompose $\bar{\mathbf{G}} = \boldsymbol\mu + \mathbf{E}$, where $\mathbf{E} = \bar{\mathbf{G}} - \boldsymbol\mu$ is the estimation error.
By matrix concentration inequalities for sums of independent random matrices~\citep{tropp2012}, the operator norm of $\mathbf{E}$ satisfies $\|\mathbf{E}\| = O(T^{-1/2})$.

Provided $\|\mathbf{E}\| < \sigma_k(\boldsymbol\mu) - \sigma_{k+1}(\boldsymbol\mu)$, Wedin's $\sin\Theta$ theorem~\citep{Wedin1972PerturbationBI} bounds the angle between the top-$k$ right singular subspaces of $\bar{\mathbf{G}}$ and $\boldsymbol{\mu}$:
\begin{equation}
    \sin\Theta\!\left(\mathcal{V}_k(\bar{\mathbf{G}}),\, \mathcal{V}_k(\boldsymbol\mu)\right) \leq \frac{\|\mathbf{E}\|}{\sigma_k(\boldsymbol\mu) - \sigma_{k+1}(\boldsymbol\mu)}.
\end{equation}
Substituting $\|\mathbf{E}\| = O(T^{-1/2})$ and using $\mathcal{V}_k(\Delta \mathbf{W}_T) = \mathcal{V}_k(\bar{\mathbf{G}})$, we obtain:
\begin{equation}
    \sin\Theta\!\left(\mathcal{V}_k(\Delta \mathbf{W}_T),\, \mathcal{V}_k(\boldsymbol\mu)\right) = O\!\left(\frac{1}{\sqrt{T}\,(\sigma_k(\boldsymbol\mu) - \sigma_{k+1}(\boldsymbol\mu))}\right).
\end{equation}
When the spectral gap $\sigma_k(\boldsymbol\mu) - \sigma_{k+1}(\boldsymbol\mu)$ is large, even a small $T$ suffices for the principal subspace of $\Delta \mathbf{W}_T$ to closely approximate that of $\boldsymbol\mu$.
This explains the rapid convergence observed in Figure~\ref{fig:method_evidence}(b): the pseudo-benign directions stabilize within 2 gradient steps because the dominant gradient directions of clean fine-tuning exhibit a substantial spectral gap.
\end{proof}

%% file: main.bbl
\begin{thebibliography}{55}
\providecommand{\natexlab}[1]{#1}
\providecommand{\url}[1]{\texttt{#1}}
\expandafter\ifx\csname urlstyle\endcsname\relax
  \providecommand{\doi}[1]{doi: #1}\else
  \providecommand{\doi}{doi: \begingroup \urlstyle{rm}\Url}\fi

\bibitem[Antol et~al.(2015)Antol, Agrawal, Lu, Mitchell, Batra, Zitnick, and
  Parikh]{vscore}
Stanislaw Antol, Aishwarya Agrawal, Jiasen Lu, Margaret Mitchell, Dhruv Batra,
  C.~Lawrence Zitnick, and Devi Parikh.
\newblock {VQA}: Visual question answering.
\newblock In \emph{ICCV}, 2015.

\bibitem[Bai et~al.(2025)Bai, Cai, Chen, Chen, Chen, Cheng, Deng, Ding, Gao,
  Ge, Ge, Guo, Huang, Huang, Huang, Hui, Jiang, Li, Li, Li, Li, Lin, Lin, Liu,
  Liu, Liu, Liu, Liu, Liu, Lu, Luo, Lv, Men, Meng, Ren, Ren, Song, Sun, Tang,
  Tu, Wan, Wang, Wang, Wang, Wang, Xie, Xu, Xu, Xu, Yang, Yang, Yang, Yang, Yu,
  Zhang, Zhang, Zhang, Zheng, Zhong, Zhou, Zhou, Zhou, Zhu, and Zhu]{qwen3vl}
Shuai Bai, Yuxuan Cai, Ruizhe Chen, Keqin Chen, Xionghui Chen, Zesen Cheng,
  Lianghao Deng, Wei Ding, Chang Gao, Chunjiang Ge, Wenbin Ge, Zhifang Guo,
  Qidong Huang, Jie Huang, Fei Huang, Binyuan Hui, Shutong Jiang, Zhaohai Li,
  Mingsheng Li, Mei Li, Kaixin Li, Zicheng Lin, Junyang Lin, Xuejing Liu,
  Jiawei Liu, Chenglong Liu, Yang Liu, Dayiheng Liu, Shixuan Liu, Dunjie Lu,
  Ruilin Luo, Chenxu Lv, Rui Men, Lingchen Meng, Xuancheng Ren, Xingzhang Ren,
  Sibo Song, Yuchong Sun, Jun Tang, Jianhong Tu, Jianqiang Wan, Peng Wang,
  Pengfei Wang, Qiuyue Wang, Yuxuan Wang, Tianbao Xie, Yiheng Xu, Haiyang Xu,
  Jin Xu, Zhibo Yang, Mingkun Yang, Jianxin Yang, An~Yang, Bowen Yu, Fei Zhang,
  Hang Zhang, Xi~Zhang, Bo~Zheng, Humen Zhong, Jingren Zhou, Fan Zhou, Jing
  Zhou, Yuanzhi Zhu, and Ke~Zhu.
\newblock {Qwen3-VL} technical report.
\newblock \emph{arXiv preprint arXiv:2511.21631}, 2025.

\bibitem[Barni et~al.(2019)Barni, Kallas, and Tondi]{sig}
Mauro Barni, Kassem Kallas, and Benedetta Tondi.
\newblock A new backdoor attack in {CNNS} by training set corruption without
  label poisoning.
\newblock In \emph{ICIP}, 2019.

\bibitem[Bj{\"o}rck \& Golub(1973)Bj{\"o}rck and Golub]{principalangles}
{\AA}ke Bj{\"o}rck and Gene~H. Golub.
\newblock Numerical methods for computing angles between linear subspaces.
\newblock \emph{Mathematics of Computation}, 1973.

\bibitem[Chen et~al.(2018)Chen, Carvalho, Baracaldo, Ludwig, Edwards, Lee,
  Molloy, and Srivastava]{activationclustering}
Bryant Chen, Wilka Carvalho, Nathalie Baracaldo, Heiko Ludwig, Benjamin
  Edwards, Taesung Lee, Ian Molloy, and Biplav Srivastava.
\newblock Detecting backdoor attacks on deep neural networks by activation
  clustering.
\newblock \emph{arXiv preprint arXiv:1811.03728}, 2018.

\bibitem[Chen et~al.(2017)Chen, Liu, Li, Lu, and Song]{blended}
Xinyun Chen, Chang Liu, Bo~Li, Kimberly Lu, and Dawn Song.
\newblock Targeted backdoor attacks on deep learning systems using data
  poisoning.
\newblock \emph{arXiv preprint arXiv:1712.05526}, 2017.

\bibitem[Chen et~al.(2020)Chen, Gong, Wang, Di, and Huang]{cloudbackdoor}
Yanjiao Chen, Xueluan Gong, Qian Wang, Xing Di, and Huayang Huang.
\newblock Backdoor attacks and defenses for deep neural networks in outsourced
  cloud environments.
\newblock \emph{IEEE Network}, 2020.

\bibitem[Dai et~al.(2023)Dai, Li, Li, Tiong, Zhao, Wang, Li, Fung, and
  Hoi]{instructblip}
Wenliang Dai, Junnan Li, Dongxu Li, Anthony Meng~Huat Tiong, Junqi Zhao,
  Weisheng Wang, Boyang Li, Pascale Fung, and Steven Hoi.
\newblock {InstructBLIP}: Towards general-purpose vision-language models with
  instruction tuning.
\newblock In \emph{NeurIPS}, 2023.

\bibitem[Gao et~al.(2025)Gao, Zhang, Zhao, Yang, and Cao]{autodrive2}
Haoxiang Gao, Li~Zhang, Yu~Zhao, Zhou Yang, and Jinghan Cao.
\newblock Application of vision-language models to pedestrian behavior
  prediction and scene understanding in autonomous driving.
\newblock In \emph{RAIIC}, 2025.

\bibitem[Gao et~al.(2019)Gao, Xu, Wang, Chen, Ranasinghe, and Nepal]{strip}
Yansong Gao, Chang Xu, Derui Wang, Shiping Chen, Damith~C. Ranasinghe, and
  Surya Nepal.
\newblock {STRIP}: a defence against trojan attacks on deep neural networks.
\newblock In \emph{ACSAC}, 2019.

\bibitem[Goyal et~al.(2017)Goyal, Khot, Summers-Stay, Batra, and Parikh]{vqav2}
Yash Goyal, Tejas Khot, Douglas Summers-Stay, Dhruv Batra, and Devi Parikh.
\newblock Making the {V} in {VQA} matter: Elevating the role of image
  understanding in visual question answering.
\newblock In \emph{CVPR}, 2017.

\bibitem[Gu et~al.(2019)Gu, Liu, Dolan-Gavitt, and Garg]{badnet}
Tianyu Gu, Kang Liu, Brendan Dolan-Gavitt, and Siddharth Garg.
\newblock {BadNets}: Evaluating backdooring attacks on deep neural networks.
\newblock \emph{IEEE Access}, 2019.

\bibitem[Guo \& Terzopoulos(2025)Guo and Terzopoulos]{medical1}
Danfeng Guo and Demetri Terzopoulos.
\newblock Prompting medical large vision-language models to diagnose
  pathologies by visual question answering.
\newblock \emph{Machine Learning for Biomedical Imaging}, 2025.

\bibitem[He et~al.(2025)He, Zhang, Liu, Lee, An, and Feng]{he2025closer}
Shuo He, Zhifang Zhang, Feng Liu, Roy Ka-Wei Lee, Bo~An, and Lei Feng.
\newblock A closer look at backdoor attacks on {CLIP}.
\newblock In \emph{ICML}, 2025.

\bibitem[Hong et~al.(2022)Hong, Carlini, and Kurakin]{handcrafted}
Sanghyun Hong, Nicholas Carlini, and Alexey Kurakin.
\newblock Handcrafted backdoors in deep neural networks.
\newblock In \emph{NeurIPS}, 2022.

\bibitem[Jiang et~al.(2026)Jiang, Liang, Rong, Zhou, Zhong, Wan, and
  Wang]{purmm}
Wenzheng Jiang, Ke~Liang, Xuankun Rong, Jingxuan Zhou, Zhengyi Zhong, Guancheng
  Wan, and Ji~Wang.
\newblock {PurMM}: Attention-guided test-time backdoor purification in
  multimodal large language models.
\newblock In \emph{AAAI}, 2026.

\bibitem[Li et~al.(2023)Li, Wong, Zhang, Usuyama, Liu, Yang, Naumann, Poon, and
  Gao]{llavamed}
Chunyuan Li, Cliff Wong, Sheng Zhang, Naoto Usuyama, Haotian Liu, Jianwei Yang,
  Tristan Naumann, Hoifung Poon, and Jianfeng Gao.
\newblock {LLaVA-Med}: Training a large language-and-vision assistant for
  biomedicine in one day.
\newblock In \emph{NeurIPS}, 2023.

\bibitem[Li et~al.(2021{\natexlab{a}})Li, Lyu, Koren, Lyu, Li, and Ma]{abl}
Yige Li, Xixiang Lyu, Nodens Koren, Lingjuan Lyu, Bo~Li, and Xingjun Ma.
\newblock Anti-backdoor learning: Training clean models on poisoned data.
\newblock In \emph{NeurIPS}, 2021{\natexlab{a}}.

\bibitem[Li et~al.(2021{\natexlab{b}})Li, Lyu, Koren, Lyu, Li, and Ma]{nad}
Yige Li, Xixiang Lyu, Nodens Koren, Lingjuan Lyu, Bo~Li, and Xingjun Ma.
\newblock Neural attention distillation: Erasing backdoor triggers from deep
  neural networks.
\newblock In \emph{ICLR}, 2021{\natexlab{b}}.

\bibitem[Li et~al.(2021{\natexlab{c}})Li, Zhai, Wu, Jiang, Li, and Xia]{st}
Yiming Li, Tongqing Zhai, Baoyuan Wu, Yong Jiang, Zhifeng Li, and Shutao Xia.
\newblock Rethinking the trigger of backdoor attack.
\newblock \emph{arXiv preprint arXiv:2004.04692}, 2021{\natexlab{c}}.

\bibitem[Li et~al.(2021{\natexlab{d}})Li, Li, Wu, Li, He, and Lyu]{issba}
Yuezun Li, Yiming Li, Baoyuan Wu, Longkang Li, Ran He, and Siwei Lyu.
\newblock Invisible backdoor attack with sample-specific triggers.
\newblock In \emph{ICCV}, 2021{\natexlab{d}}.

\bibitem[Liang et~al.(2025)Liang, Liang, Pang, Du, Liu, Zhu, Cao, and
  Tao]{revisiting_lvlm_backdoor}
Siyuan Liang, Jiawei Liang, Tianyu Pang, Chao Du, Aishan Liu, Mingli Zhu,
  Xiaochun Cao, and Dacheng Tao.
\newblock Revisiting backdoor attacks against large vision-language models from
  domain shift.
\newblock In \emph{CVPR}, 2025.

\bibitem[Lin et~al.(2014)Lin, Maire, Belongie, Hays, Perona, Ramanan,
  Doll{\'{a}}r, and Zitnick]{coco}
Tsung{-}Yi Lin, Michael Maire, Serge~J. Belongie, James Hays, Pietro Perona,
  Deva Ramanan, Piotr Doll{\'{a}}r, and C.~Lawrence Zitnick.
\newblock Microsoft {COCO:} common objects in context.
\newblock In \emph{ECCV}, 2014.

\bibitem[Lin et~al.(2024)Lin, Liu, Wei, Li, and Xiong]{tsbd}
Weilin Lin, Li~Liu, Shaokui Wei, Jianze Li, and Hui Xiong.
\newblock Unveiling and mitigating backdoor vulnerabilities based on unlearning
  weight changes and backdoor activeness.
\newblock In \emph{NeurIPS}, 2024.

\bibitem[Liu et~al.(2024)Liu, Li, Li, and Lee]{llava}
Haotian Liu, Chunyuan Li, Yuheng Li, and Yong~Jae Lee.
\newblock Improved baselines with visual instruction tuning.
\newblock In \emph{CVPR}, 2024.

\bibitem[Liu et~al.(2018)Liu, Dolan-Gavitt, and Garg]{finepruning}
Kang Liu, Brendan Dolan-Gavitt, and Siddharth Garg.
\newblock {Fine-Pruning}: Defending against backdooring attacks on deep neural
  networks.
\newblock In \emph{RAID}, 2018.

\bibitem[Liu \& Zhang(2025)Liu and Zhang]{badvision}
Zhaoyi Liu and Huan Zhang.
\newblock Stealthy backdoor attack in self-supervised learning vision encoders
  for large vision language models.
\newblock In \emph{CVPR}, 2025.

\bibitem[Lyu et~al.(2024)Lyu, Pang, Ma, Ling, and Chen]{trojvlm}
Weimin Lyu, Lu~Pang, Tengfei Ma, Haibin Ling, and Chao Chen.
\newblock {TrojVLM}: Backdoor attack against vision language models.
\newblock In \emph{ECCV}, 2024.

\bibitem[Lyu et~al.(2025)Lyu, Yao, Gupta, Pang, Sun, Yi, Hu, Ling, and
  Chen]{vlood}
Weimin Lyu, Jiachen Yao, Saumya Gupta, Lu~Pang, Tao Sun, Lingjie Yi, Lijie Hu,
  Haibin Ling, and Chao Chen.
\newblock Backdooring vision-language models with out-of-distribution data.
\newblock In \emph{ICLR}, 2025.

\bibitem[Ma et~al.(2025)Ma, Gao, Wang, Wang, Wang, Sun, Ding, Xu, Chen, Zhao,
  Huang, Li, Wu, Zhang, Zheng, Bai, Li, Wu, Qiu, Zhang, Han, Li, Sun, Wang, Gu,
  Wu, Chen, Zhang, Liu, Gong, Liu, Pan, Xie, Pang, Dong, Jia, Zhang, Ma, Zhang,
  Gong, Xiao, Erfani, Baldwin, Li, Sugiyama, Tao, Bailey, and
  Jiang]{safetysurvey1}
Xingjun Ma, Yifeng Gao, Yixu Wang, Ruofan Wang, Xin Wang, Ye~Sun, Yifan Ding,
  Hengyuan Xu, Yunhao Chen, Yunhan Zhao, Hanxun Huang, Yige Li, Yutao Wu,
  Jiaming Zhang, Xiang Zheng, Yang Bai, Yiming Li, Zuxuan Wu, Xipeng Qiu,
  Jingfeng Zhang, Xudong Han, Haonan Li, Jun Sun, Cong Wang, Jindong Gu,
  Baoyuan Wu, Siheng Chen, Tianwei Zhang, Yang Liu, Mingming Gong, Tongliang
  Liu, Shirui Pan, Cihang Xie, Tianyu Pang, Yinpeng Dong, Ruoxi Jia, Yang
  Zhang, Shiqing Ma, Xiangyu Zhang, Neil Gong, Chaowei Xiao, Sarah Erfani, Tim
  Baldwin, Bo~Li, Masashi Sugiyama, Dacheng Tao, James Bailey, and Yu-Gang
  Jiang.
\newblock Safety at scale: a comprehensive survey of large model and agent
  safety.
\newblock \emph{Foundations and Trends in Privacy and Security}, 2025.

\bibitem[Min et~al.(2025)Min, Pham, and Sun]{ulrl}
Nay~Myat Min, Long~H. Pham, and Jun Sun.
\newblock Unified neural backdoor removal with only few clean samples through
  unlearning and relearning.
\newblock \emph{T-IFS}, 2025.

\bibitem[Nguyen \& Tran(2020)Nguyen and Tran]{inputaware}
Tuan~Anh Nguyen and Anh Tran.
\newblock Input-aware dynamic backdoor attack.
\newblock In \emph{NeurIPS}, 2020.

\bibitem[Nguyen \& Tran(2021)Nguyen and Tran]{wanet}
Tuan~Anh Nguyen and Anh~Tuan Tran.
\newblock {WaNet} - imperceptible warping-based backdoor attack.
\newblock In \emph{ICLR}, 2021.

\bibitem[Rong et~al.(2025)Rong, Huang, Liang, Bi, Xiao, Li, Du, and Ye]{bye}
Xuankun Rong, Wenke Huang, Jian Liang, Jinhe Bi, Xun Xiao, Yiming Li, Bo~Du,
  and Mang Ye.
\newblock Backdoor cleaning without external guidance in {MLLM} fine-tuning.
\newblock In \emph{NeurIPS}, 2025.

\bibitem[Shi et~al.(2023)Shi, Du, Wu, Guan, Sun, and Liu]{zip}
Yucheng Shi, Mengnan Du, Xuansheng Wu, Zihan Guan, Jin Sun, and Ninghao Liu.
\newblock Black-box backdoor defense via zero-shot image purification.
\newblock In \emph{NeurIPS}, 2023.

\bibitem[Tran et~al.(2018)Tran, Li, and Madry]{spectralsignatures}
Brandon Tran, Jerry Li, and Aleksander Madry.
\newblock Spectral signatures in backdoor attacks.
\newblock In \emph{NeurIPS}, 2018.

\bibitem[Tropp(2012)]{tropp2012}
Joel~A. Tropp.
\newblock User-friendly tail bounds for sums of random matrices.
\newblock \emph{Found. Comput. Math.}, 2012.

\bibitem[Van et~al.(2024)Van, Verma, and Wu]{medical2}
Minh–Hao Van, Prateek Verma, and Xintao Wu.
\newblock On large visual language models for medical imaging analysis: An
  empirical study.
\newblock In \emph{CHASE}, 2024.

\bibitem[Vedantam et~al.(2015)Vedantam, Zitnick, and Parikh]{cider}
Ramakrishna Vedantam, C.~Lawrence Zitnick, and Devi Parikh.
\newblock {CIDEr}: Consensus-based image description evaluation.
\newblock In \emph{CVPR}, 2015.

\bibitem[Wang et~al.(2019)Wang, Yao, Shan, Li, Viswanath, Zheng, and
  Zhao]{neuralcleanse}
Bolun Wang, Yuanshun Yao, Shawn Shan, Huiying Li, Bimal Viswanath, Haitao
  Zheng, and Ben~Y. Zhao.
\newblock {Neural Cleanse}: Identifying and mitigating backdoor attacks in
  neural networks.
\newblock In \emph{IEEE S\&P}, 2019.

\bibitem[Wang et~al.(2025)Wang, Zhu, Jin, Chen, and Lim]{lpp}
Xinyi Wang, Zhiyu Zhu, Zhibo Jin, Huaming Chen, and Teng~Joon Lim.
\newblock Rethinking {Lipschitzness} data-free backdoor defense.
\newblock In \emph{CIKM}, 2025.

\bibitem[Wedin(1972)]{Wedin1972PerturbationBI}
Per-\r{A}ke Wedin.
\newblock Perturbation bounds in connection with singular value decomposition.
\newblock \emph{BIT}, 1972.

\bibitem[Wu \& Wang(2021)Wu and Wang]{anp}
Dongxian Wu and Yisen Wang.
\newblock Adversarial neuron pruning purifies backdoored deep models.
\newblock In \emph{NeurIPS}, 2021.

\bibitem[Xu et~al.(2025)Xu, Liang, Zheng, Liu, Wang, Luo, Lin, Rutkowski, and
  Tao]{srd}
Shuhan Xu, Siyuan Liang, Hongling Zheng, Aishan Liu, Xinbiao Wang, Yong Luo,
  Fu~Lin, Leszek Rutkowski, and Dacheng Tao.
\newblock {SRD}: Reinforcement-learned semantic perturbation for backdoor
  defense in {VLMs}.
\newblock \emph{arXiv preprint arXiv:2506.04743}, 2025.

\bibitem[Ye et~al.(2025)Ye, Rong, Huang, Du, Yu, and Tao]{safetysurvey2}
Mang Ye, Xuankun Rong, Wenke Huang, Bo~Du, Nenghai Yu, and Dacheng Tao.
\newblock A survey of safety on large vision-language models: Attacks, defenses
  and evaluations.
\newblock \emph{arXiv preprint arXiv:2502.14881}, 2025.

\bibitem[Ye et~al.(2024)Ye, Xu, Ye, Yan, Hu, Liu, Qian, Zhang, and
  Huang]{mplugowl2}
Qinghao Ye, Haiyang Xu, Jiabo Ye, Ming Yan, Anwen Hu, Haowei Liu, Qi~Qian,
  Ji~Zhang, and Fei Huang.
\newblock {mPLUG-Owl2}: Revolutionizing multi-modal large language model with
  modality collaboration.
\newblock In \emph{CVPR}, 2024.

\bibitem[Zeng et~al.(2022)Zeng, Chen, Park, Mao, Jin, and Jia]{ibau}
Yi~Zeng, Si~Chen, Won Park, Zhuoqing Mao, Ming Jin, and Ruoxi Jia.
\newblock Adversarial unlearning of backdoors via implicit hypergradient.
\newblock In \emph{ICLR}, 2022.

\bibitem[Zhang et~al.(2025)Zhang, He, Wang, Shen, and Feng]{zhang2024defending}
Zhifang Zhang, Shuo He, Haobo Wang, Bingquan Shen, and Lei Feng.
\newblock Defending multimodal backdoored models by repulsive visual prompt
  tuning.
\newblock In \emph{NeurIPS}, 2025.

\bibitem[Zhang et~al.(2026{\natexlab{a}})Zhang, Tao, Lv, Zhao, Feng, and
  Zhou]{zhang2025tokenswap}
Zhifang Zhang, Qiqi Tao, Jiaqi Lv, Na~Zhao, Lei Feng, and Joey~Tianyi Zhou.
\newblock {TokenSwap}: Backdoor attack on the compositional understanding of
  large vision-language models.
\newblock In \emph{ICML}, 2026{\natexlab{a}}.

\bibitem[Zhang et~al.(2026{\natexlab{b}})Zhang, Yang, He, Chen, Zhang, Maennel,
  Feng, and Xu]{cleansight}
Zhifang Zhang, Bojun Yang, Shuo He, Weitong Chen, Wei~Emma Zhang, Olaf Maennel,
  Lei Feng, and Miao Xu.
\newblock Test-time attention purification for backdoored large vision language
  models.
\newblock In \emph{CVPR}, 2026{\natexlab{b}}.

\bibitem[Zhang et~al.(2026{\natexlab{c}})Zhang, Zhang, Zhou, Wei, He, Liu, and
  Feng]{zhang2025improving}
Zhifang Zhang, Jiahan Zhang, Shengjie Zhou, Qi~Wei, Shuo He, Feng Liu, and Lei
  Feng.
\newblock Improving generalizability and undetectability for targeted
  adversarial attacks on multimodal pre-trained models.
\newblock \emph{TPAMI}, 2026{\natexlab{c}}.

\bibitem[Zhao et~al.(2025)Zhao, Zhu, Jiao, Gao, and Wu]{safetysurvey3}
Pinlong Zhao, Weiyao Zhu, Pengfei Jiao, Di~Gao, and Ou~Wu.
\newblock Data poisoning in deep learning: {A} survey.
\newblock \emph{arXiv preprint arXiv:2503.22759}, 2025.

\bibitem[Zheng et~al.(2022)Zheng, Tang, Li, and Liu]{clp}
Runkai Zheng, Rongjun Tang, Jianze Li, and Li~Liu.
\newblock Data-free backdoor removal based on channel {Lipschitzness}.
\newblock In \emph{ECCV}, 2022.

\bibitem[Zhou et~al.(2024)Zhou, Liu, Yurtsever, Zagar, Zimmer, Cao, and
  Knoll]{autodriving}
Xingcheng Zhou, Mingyu Liu, Ekim Yurtsever, Bare~Luka Zagar, Walter Zimmer,
  Hu~Cao, and Alois~C. Knoll.
\newblock Vision language models in autonomous driving: A survey and outlook.
\newblock \emph{T-IV}, 2024.

\bibitem[Zhu et~al.(2024)Zhu, Chen, Shen, Li, and Elhoseiny]{minigpt4}
Deyao Zhu, Jun Chen, Xiaoqian Shen, Xiang Li, and Mohamed Elhoseiny.
\newblock {MiniGPT-4}: Enhancing vision-language understanding with advanced
  large language models.
\newblock In \emph{ICLR}, 2024.

\end{thebibliography}
